%% file: main.tex
\documentclass[acmtog]{acmart} 

\usepackage{tabularx}
\usepackage{censor}

\usepackage{tcolorbox}

\newcommand{\lm}[1]{{\color{black}#1}}

\AtBeginDocument{%
  }

\copyrightyear{2026}
\acmYear{2026}
\setcopyright{cc}
\setcctype{by}
\acmConference[SIGGRAPH Conference Papers '26]{Special Interest Group on Computer Graphics and Interactive Techniques Conference Conference Papers}{July 19--23, 2026}{Los Angeles, CA, USA}
\acmBooktitle{Special Interest Group on Computer Graphics and Interactive Techniques Conference Conference Papers (SIGGRAPH Conference Papers '26), July 19--23, 2026, Los Angeles, CA, USA}
\acmDOI{10.1145/3799902.3811082}
\acmISBN{979-8-4007-2554-8/2026/07}

\setcitestyle{authoryear}
\usepackage{cleveref}
\usepackage{multirow}
\usepackage{algpseudocode}

\begin{document}

\title{Abstraction in Style: Beyond Texture and Color}

\author{Min Lu}
\email{lumin.vis@gmail.com}
\affiliation{%
  \institution{Shenzhen University}
  \country{China}
}

\author{Yuanfeng He}
\affiliation{%
\department{Visual Computing Research Center (VCC), College of Computer Science and Software Engineering (CSSE)}
  \institution{Shenzhen University}
  \country{China}
}

\author{Anthony Chen}
\affiliation{%
  \institution{Peking University}
  \country{China}
}

\author{Jianhuang He}
\affiliation{%
\department{VCC, CSSE}
  \institution{Shenzhen University}
  \country{China}
}

\author{Pu Wang}
\affiliation{%
  \institution{Shenzhen University}
  \country{China}
}

\author{Daniel Cohen-Or}
\affiliation{%
\department{VCC, CSSE}
  \institution{Tel Aviv University}
  \country{Israel,}
  \institution{Shenzhen University}
  \country{China}
}

\author{Hui Huang}
\affiliation{%
	\department{VCC, CSSE}
	\institution{Shenzhen University}
	\country{China}	
}
\email{hhzhiyan@gmail.com}
\authornote{Corresponding author: Hui Huang (hhzhiyan@gmail.com)}


\renewcommand{\shortauthors}{M. Lu, Y. He, A. Chen, J. Huang, P. Wu, D. Cohen-Or, and H. Huang}


\begin{abstract}   
\input{tex/abstract}
\end{abstract}

\begin{teaserfigure}
  \includegraphics[width=\textwidth]{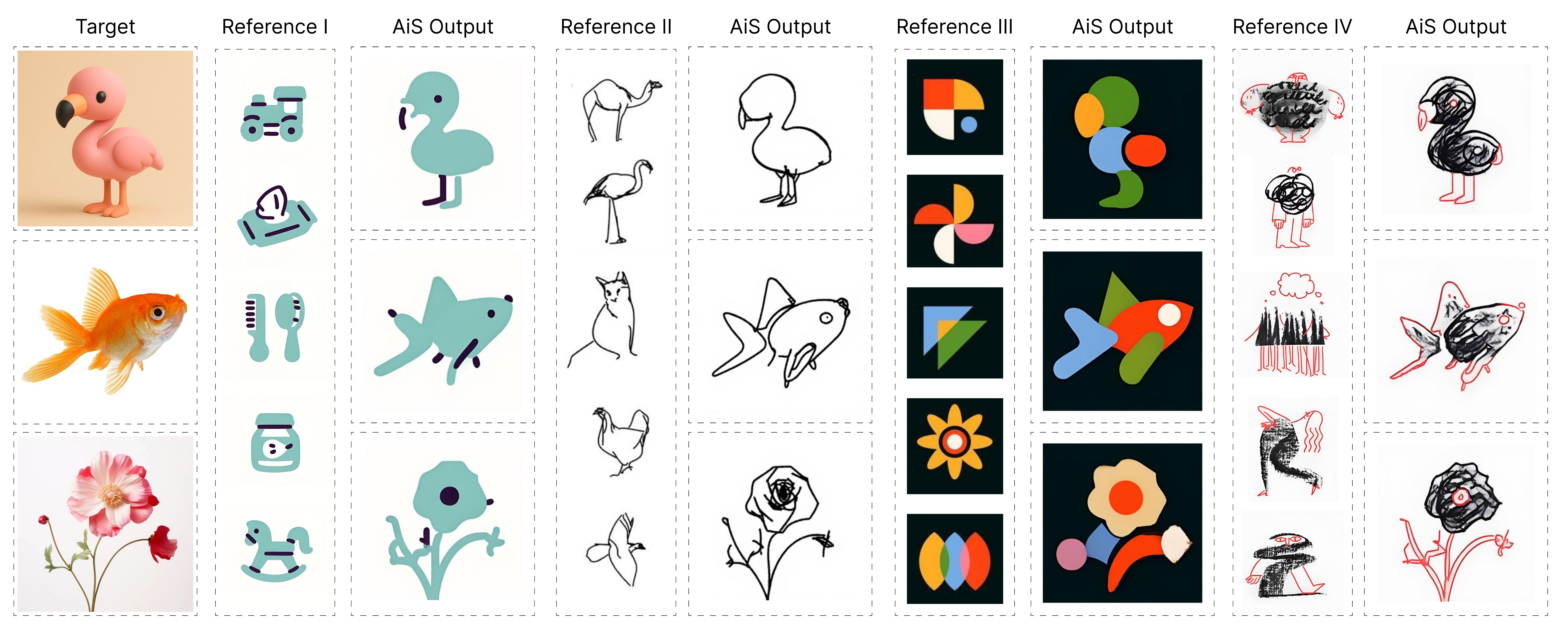}
  \caption{\textbf{Abstraction in Style (AiS):} Across four challenging illustrative styles with strong structural abstraction, AiS generates coherent results for target images from a small set of exemplars, extending style transfer beyond color and texture to include geometric and structural deformations. Reference images sourced from Pinterest; \copyright~ original authors.}
  \label{fig:teaser}
\end{teaserfigure}





\maketitle

\input{tex/1_intro}

\input{tex/2_related-works}

\input{tex/3_method}

\input{tex/32_VisualStyle}

\input{tex/4_VAT}

\input{tex/5_experiments}
\input{tex/discussion}
\begin{acks}
This work was supported in part by ICFCRT (W2441020), NSFC (62472288), Guangdong Basic and Applied Basic Research Foundation (2023B1515120026), Shenzhen Science and Technology Program (KQTD20210811090044003), ISF (2492/20, 1473/24), MNR Key Laboratory for Geo-Environmental Monitoring of Great Bay Area, Guangdong Key Laboratory of Urban Informatics, and State Key Laboratory of Subtropical Building and Urban Science.
\end{acks}

\bibliographystyle{ACM-Reference-Format}
\bibliography{references}


\input{tex/figures_only}

\clearpage

\appendix


\input{tex/supp_content}

\clearpage

\input{tex/figure_only_supp}

\end{document}

%% file: tex/abstract.tex
Artistic styles often embed abstraction beyond surface appearance, involving deliberate reinterpretation of structure rather than mere changes in texture or color.
Conventional style transfer methods typically preserve the input geometry and therefore struggle to capture this deeper abstraction behavior, especially for illustrative and non-photorealistic styles.
In this work, we introduce \emph{Abstraction in Style (AiS)}, a generative framework that 
separates \emph{structural abstraction} from \emph{visual stylization}.
Given a target image and a small set of style exemplars, AiS first derives an intermediate \emph{abstraction proxy} that reinterprets the target’s structure in accordance with the abstraction logic exhibited by the style.
The proxy captures semantic structure while relaxing geometric fidelity, enabling subsequent stylization to operate on an abstracted representation rather than the original image.
In a second stage, the abstraction proxy is rendered to produce the final stylized output, preserving visual coherence with the reference style.
Both stages are implemented using a shared image-space analogy, enabling transformations to be learned from visual exemplars without explicit geometric supervision.
By decoupling abstraction from appearance and treating abstraction as an explicit, transferable process, AiS supports a wider range of stylistic transformations, improves controllability, and enables more expressive stylization.



%% file: tex/1_intro.tex
\section{Introduction}

The generation of stylized visual content is a longstanding goal in computer graphics~\cite{gatys2016image}.
A substantial body of research is dedicated to style transfer, in which the visual traits of a reference artwork, such as stroke texture, color palette, or ornamentation, are applied to a target image while preserving its general underlying subject. Many works transfer style in a way that adheres to the original spatial structure~\cite{AttDistill25, wang2024instantstyle},
even when the target style is abstract, and therefore tend to overlook the abstraction function that may be latent in the reference style. 

\begin{figure}[!htb]
  \centering    \includegraphics[width=1.\linewidth]{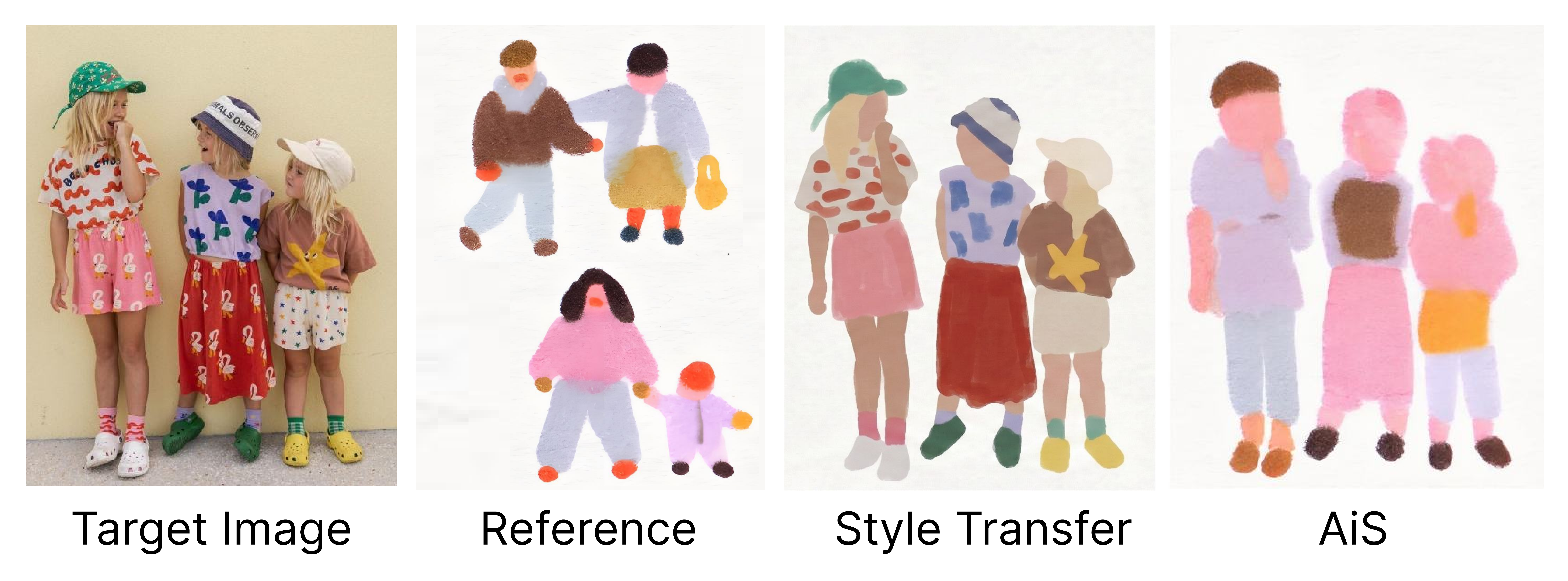}
            \caption{Conventional style transfer keeps the structure of the target image. Introducing an abstraction stage prior to the stylization, allows adopting structural attributes from the reference, leading to a more coherent transformation.}
  \label{fig:quick_example}
\end{figure}

In this paper, we introduce \emph{Abstraction in Style (AiS)}, a generative framework grounded in the premise that style and abstraction are distinct yet intertwined dimensions of visual expression (see the references in Figure~\ref{fig:teaser}). \lm{While style governs surface appearance like texture and color, abstraction involves a reinterpretation of structure, focusing on re-expressing essential meaning or relationships within forms rather than replicating exact, realistic geometry.} \lm{Abstraction that alters the geometric form of a subject~\cite{hale2012abstraction} is central to illustrative and non-photorealistic styles}: lines may become deliberately irregular~\cite{yaniv2019face}, less significant geometric features can be smoothed away~\cite{mehra2009abstraction, Grabler2008_touristmap}, symmetry may be intentionally broken, and proportions distorted to convey a particular visual character or simplification. These operations reflect a higher-level visual reasoning that conventional style transfer methods, which largely preserve input geometry, are not designed to capture. AiS addresses this limitation by explicitly separating \emph{structural abstraction} from \emph{visual stylization}, enabling the model not only to transfer visual appearance but also to capture the abstraction behavior inherent in a reference style, defined by a small set of exemplars. As illustrated in Figure~\ref{fig:quick_example}, this explicit treatment of abstraction allows structural reinterpretation beyond rigid geometry preservation, which is essential for many illustrative and non-photorealistic styles \lm{that AiS aims to capture}.


Conceptually, AiS formulates stylized image generation as a two-stage process that separates structural abstraction from visual stylization. In the first stage, \emph{Structural Abstraction}, the structure of the target image is reinterpreted in the spirit of the reference exemplars, producing an intermediate \emph{abstraction proxy} that reflects the abstraction tendencies in the references while preserving the subject identity of the target. In the second stage, \emph{Visual Stylization}, this proxy is rendered into a final image that is visually coherent with the reference style. 

In this paper, we present one concrete realization of this framework \lm{tailored for object-centric images}. Taking the \emph{abstraction proxy} as a bridge, both the abstraction and stylization stages are implemented using a shared image-space analogy mechanism, referred to as \emph{Visual Analogy Transfer (VAT)}. 
Trained on a small set of style exemplars (averagely 10 per style), VAT learns transformations directly from these images without explicit geometric supervision. 
While alternative realizations of AiS are possible, our implementation demonstrates that explicitly modeling abstraction alongside appearance yields a more interpretable and controllable stylization process and supports a broader range of structural transformations than conventional style transfer methods.


We demonstrate the effectiveness of Abstraction in Style through extensive qualitative results and evaluations. As shown in Figure~\ref{fig:teaser}, AiS supports stylized image generation across diverse illustrative styles, producing outputs that reflect both the visual appearance and the underlying abstraction logic of the reference. By explicitly modeling abstraction as a separate stage, AiS yields more coherent structural reinterpretations than conventional style transfer methods, particularly for non-photorealistic and highly abstract styles. Comparative experiments, ablations, and user studies further validate that disentangling abstraction from appearance improves stylistic fidelity, structural expressiveness, and controllability.

%% file: tex/2_related-works.tex
\section{Related Work}

Prior work related to Abstraction in Style spans two largely independent
directions: visual abstraction, which reinterprets structure, and style
transfer, which modifies appearance while typically preserving geometry.
While both have been extensively studied, they have largely evolved
separately, without explicitly modeling abstraction as a transferable
process conditioned on style exemplars.

Visual abstraction has long been a core concept in art and design
\cite{hale2012abstraction}. Early computational approaches explored
geometry-driven abstractions through skeletal representations, part-based
models, planar slicing, and shape proxies
\cite{de2011exoskeleton,demirci2008skeletal,mccrae2011slices,mi2009abstraction,sala2010contour,mehra2009abstraction,nan2011conjoining,yumer2012co}.
\lm{In imagery, abstraction is closely tied to non-photorealistic rendering
\cite{decarlo2002stylization}, with seminal work on painterly and
brush-based representations
\cite{haeberli1990paint,hertzmann1998painterly,litwinowicz1997processing,shiraishi2000algorithm,gooch2002_artisticvision, Song2008_artyshapes, Zeng2009_painterlyrendering}. While some approaches, such as Gooch's artistic vision~\cite{gooch2002_artisticvision}, use similar concepts of skeletonization, they rely on redrawing with pre-defined brushes or shapes rather than exemplar-driven learning.} Sketch-based abstraction has emerged as a particularly expressive domain,
including data-driven stroke reassembly and learning-based sketchy stroke pruning
approaches
\cite{berger2013style,muhammad2018learning,liu2021neural}. \lm{More recent optimization-based methods leverage pretrained vision–language
models to achieve semantic sketchy representations \cite{vinker2022clipasso,vinker2023clipascene}, but these approaches typically operate in a fixed or implicit style and do not condition abstraction behavior on exemplars.}

Style transfer aims to modify visual appearance while preserving underlying
structure, most notably introduced by ~\citeN{gatys2016image}.
Subsequent work improved efficiency and flexibility through feed-forward
networks, feature normalization, and universal style representations
\cite{johnson2016perceptual,ulyanov2016texture,ulyanov2017improved,huang2017arbitrary,chen2016fast,li2017universal,zhang2023inversion}.
Recent diffusion-based methods enable exemplar-based stylization by
modulating attention or adapting pretrained models with lightweight
modules, allowing consistent appearance transfer across generations
\cite{alaluf2023crossimage,hertz2024stylealigned,chung2024style,frenkel2024blora,shah2024ziplora,ouyang2025k,roy2025duolora}.
Despite their effectiveness, these methods largely preserve the input
geometry and do not explicitly model the abstraction behavior inherent in
many illustrative styles.

In contrast to prior work, Abstraction in Style explicitly separates
structural abstraction from visual stylization and treats abstraction as a
learnable, transferable process conditioned on style exemplars. This
decoupling enables structural reinterpretation beyond rigid geometry
preservation, while retaining compatibility with exemplar-driven
stylization frameworks.

%% file: tex/3_method.tex

\vspace{-1.0em}
\section{Overview}

Abstraction in Style (AiS) formulates stylized image generation as a process that explicitly separates \emph{structural abstraction} from \emph{visual stylization}.
Many \lm{illustrative and non-photorealistic} styles are encoded not merely through appearance, but through deliberate reinterpretation of structure, like simplifying, exaggerating, or reshaping geometry to convey semantic or perceptual intent.
\lm{Conventional style transfer methods, largely constrained by rigid geometry preservation, struggle to capture this deeper abstraction logic, particularly in the highly abstract regimes characteristic of illustrative art}. 

\begin{figure}[!t]
  \centering    \includegraphics[width=1.\linewidth]{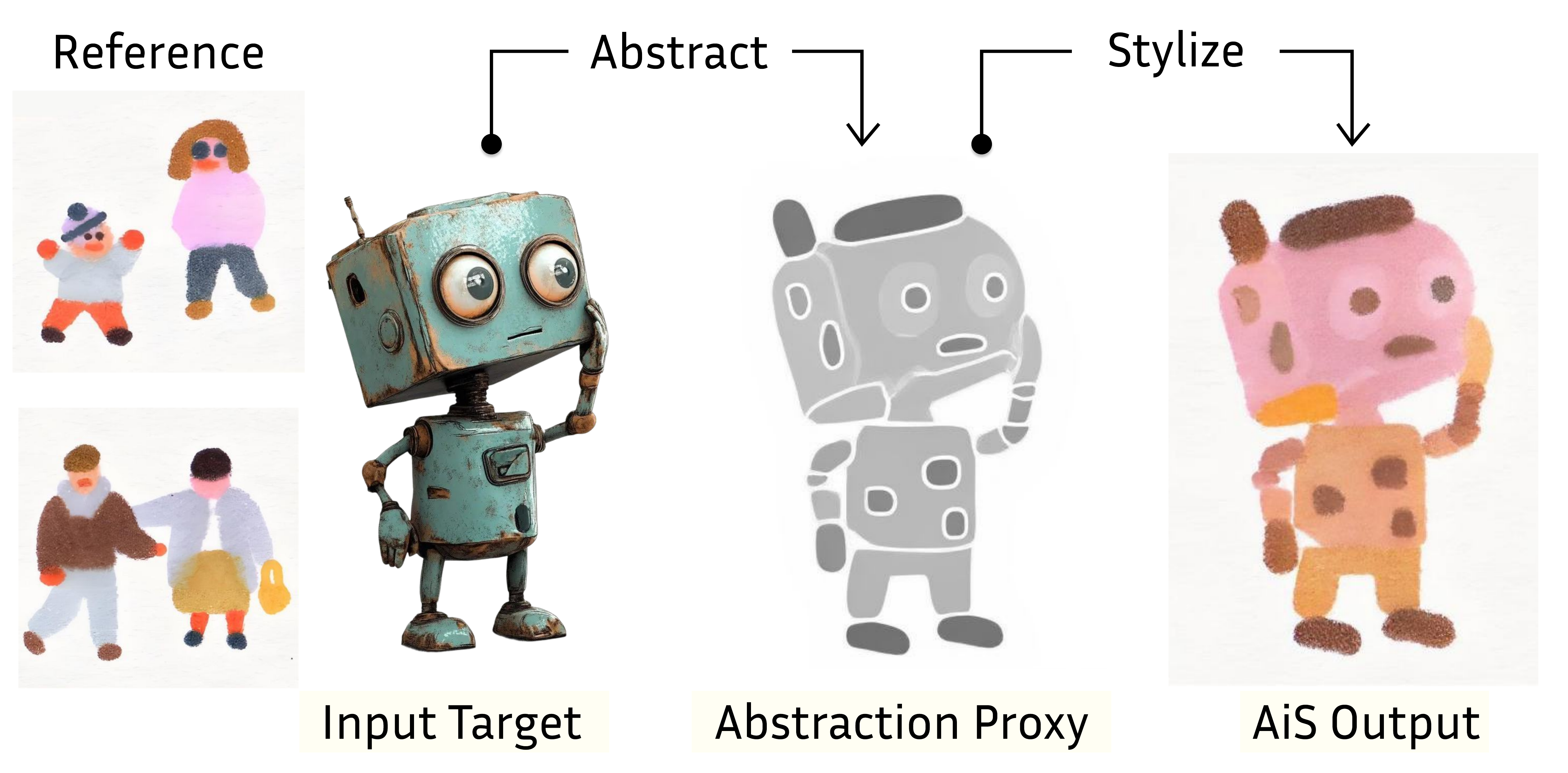}
    \caption{Two stages of AiS: in the abstraction stage, the target image is reshaped following the reference, generating an \textit{abstraction proxy}; Then, in the second stage, the proxy is stylized to produce the final output.}
  \label{fig:two_stages}
\end{figure}

To address this, AiS decomposes stylized image generation into two sequential stages (Figure~\ref {fig:two_stages}).
In the first stage, \emph{Structural Abstraction}, the structure of the target image is reinterpreted in the spirit of the reference exemplars, producing an intermediate representation (called \textit{Abstraction Proxy}) that reflects the abstraction tendencies of the reference.
In the second stage, \emph{Visual Stylization}, this abstracted structure is rendered in a manner consistent with the visual appearance of the exemplars.
This separation provides a conceptual framework for disentangling abstraction from appearance, transforming stylization from a single monolithic operation into a more interpretable and controllable process.

Underlying both stages of our framework is a shared analogy-based formulation ($A \rightarrow A' :: B \rightarrow B'$). Specifically, in both Structural Abstraction and Visual Stylization, we formulate the transformation between representations as a visual analogy learning problem in image space.
This formulation, which we refer to as \textit{Visual Analogy Transfer} (VAT), enables the foundation generative model to learn transformations from reference exemplars and apply them to the input target entirely in image space, bypassing the need for explicit geometric operations or handcrafted features.

Within AiS, this formulation is instantiated in two distinct settings.
In the abstraction stage, \emph{Abstraction VAT (A-VAT)} learns how structural representations are reinterpreted according to the abstraction logic of given references.
In the stylization stage, \emph{Stylization VAT (S-VAT)} applies the same analogy principle to learn the appearance transformation and render an abstraction proxy into a fully stylized image.
Although both follow the same analogy-based formulation, they operate over different representations and serve different roles within the pipeline.
The following sections describe A-VAT and S-VAT in detail, while Section~6 presents the general VAT formulation and its implementation.

\section{Structural Abstraction}
\label{sec:structural_abstraction}

Given the target image as input, the structural abstraction stage produces its \textit{Abstraction Proxy}, an intermediate image that explicitly represents how the target’s structure is reinterpreted under the abstraction logic of the references. Serving as a bridge between the input content and the final stylized output, this abstraction proxy 
captures which elements are present and how parts relate, without committing to a particular geometric realization.

\begin{figure}[!h]
  \centering    \includegraphics[width=1.\linewidth]{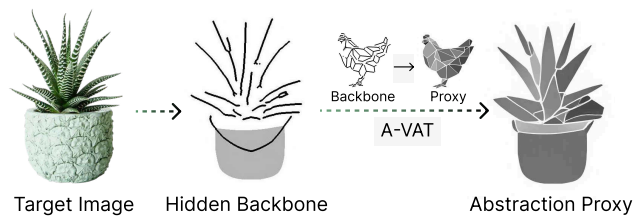}
    \caption{Structural Abstraction Process: given a target image, its hidden backbone is first computed via skeleton detection and area erosion. An abstraction proxy is then generated by the A-VAT, which is trained to learn the visual analogy from the Backbone $\rightarrow$ Proxy pairs in the reference set, e.g., the pair of roosters.} 
  \label{fig:abstraction}
\end{figure}

To generate the abstraction proxy for the given target, the abstraction stage proceeds in two steps, as shown in Figure~\ref{fig:abstraction}. 
First, the target is converted into a \emph{Hidden Backbone} image (hereafter \textit{Backbone}), a simplified structural representation that deliberately strips away visual appearance and degrades geometry while preserving the object's semantic layout and topological organization. The backbone serves as a style-agnostic structural representation of the shape, providing a common foundation for abstraction.
In the second step, the hidden backbone is transformed into an abstraction proxy using the trained A-VAT, which accomplishes the following visual analogy:

\begin{equation*}
    \text{Backbone}_r \rightarrow \text{Proxy}_r  ::  \text{Backbone}_t \rightarrow \text{Proxy}_t
\end{equation*}
where the A-VAT learns the transformation from hidden backbone to abstraction proxy from reference pairs \textit{$(\text{Backbone}_r \rightarrow \text{Proxy}_r)$}, then applies it to synthesize the abstraction proxy ($\text{Proxy}_t$) given the target backbone ($\text{Backbone}_t$). 

Rather than applying fixed geometric rules or heuristic transformation goals, the A-VAT learns how the structure is abstracted by observing the image pairs of \textit{Backbone $\rightarrow$ Proxy} in the references and applies the same abstraction logic to the target. 
Importantly, it should be noted that the A-VAT operates entirely in image space and does not rely on explicit geometric reconstruction.

\subsection{Construction of Hidden Backbone}
\label{subsec:hidden_struct}

\begin{figure}[!htb]
  \centering    \includegraphics[width=1.\linewidth]{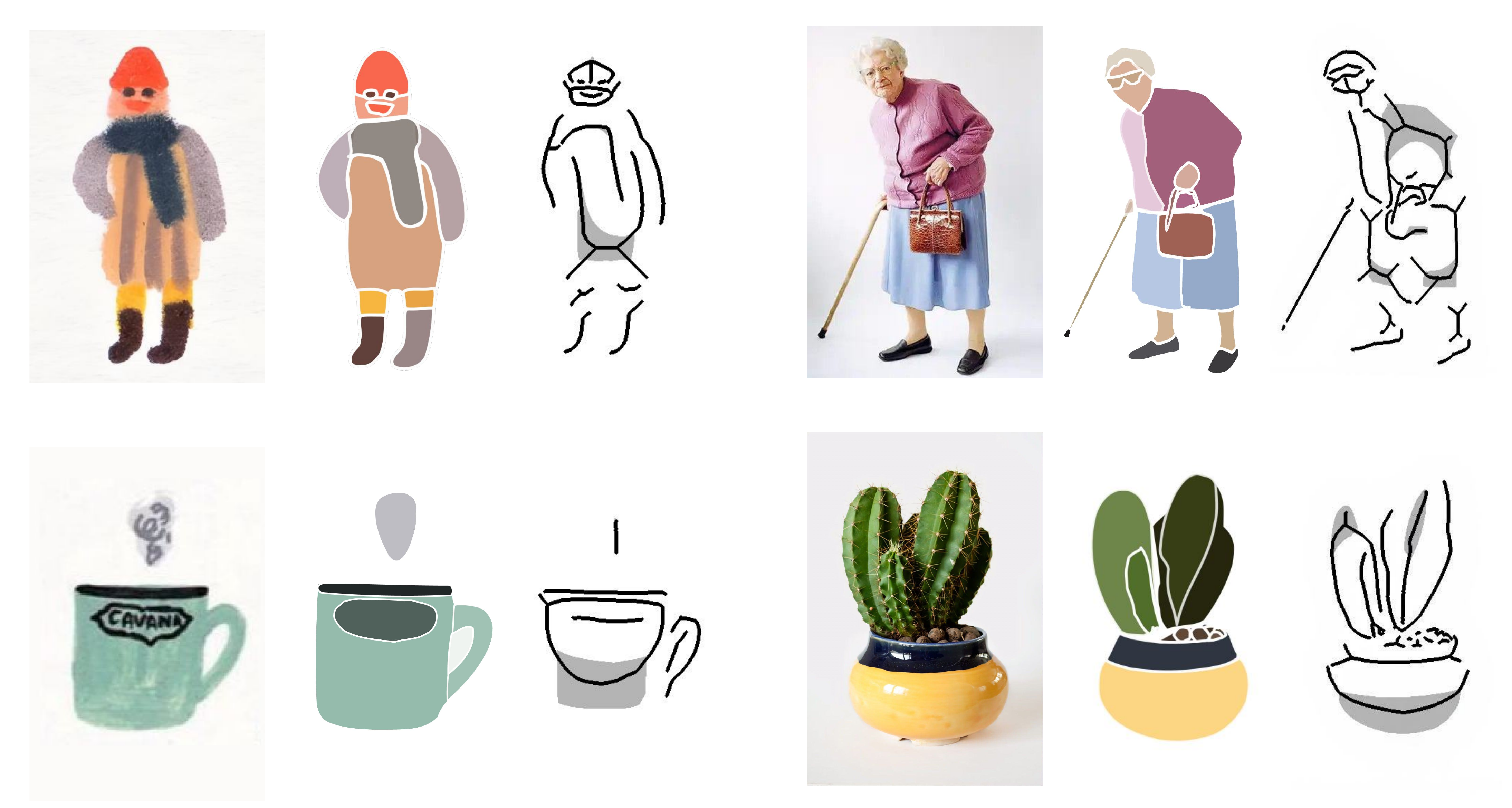}
    \caption{Hidden Backbone Construction: four examples are shown. For each example, the process proceeds as follows: the input target image (left) is first vectorized into simplified flat-color shapes (middle); the skeleton is then detected, and regions are eroded to yield the final hidden structure (right).} 
  \label{fig:hidden_stru_form}
\end{figure}

The idea behind \textit{Hidden Backbone} is a representation that distills precise boundaries into a structural representation, retaining only the visual essence necessary to identify the subject. This simplified backbone provides a flexible scaffold for geometric reimagination in the A-VAT. \lm{While other representations, such as segmentation-based tokens~\cite{gooch2002_artisticvision, Zeng2009_painterlyrendering}, can also capture image structure, we adopt the skeleton combined with eroded areas as the hidden backbone in this work. This choice is motivated by its conceptual simplicity and general applicability across object categories.} The construction of the hidden structure is illustrated in Figure~\ref{fig:hidden_stru_form}, which involves two steps: first, extraction of simplified areas, and then skeletonization and region erosion.

\paragraph{Vectorization for Simplified Shapes} \lm{To remove fine-grained visual complexity, we first process the input image, which often contains intricate details and textures, into a region-based structural representation. Specifically, we adopt the layered image vectorization method of \citet{wang2024layered} to convert regions of consistent appearance into clean vector paths with flat colors. This step effectively eliminates photographic noise and provides a simplified shape representation that serves as the foundation for our backbone analysis.}

\paragraph{Skeleton Extraction \& Area Erosion} With shapes simplified through vectorization, we rasterize them into binary images. We adopt a medial axis-based approach, where the topological backbone is first extracted via morphological skeletonization, which iteratively peels away boundary pixels to yield a two-pixel-wide centerline. \lm{While alternative simplification methods exist, for instance based on Gestalt principles~\cite{nan2011conjoining, barla2006stroke, liu2015closure}, this morphological variant suffices for our current purposes; exploring sophisticated alternatives remains a subject for future work.}
However, a skeleton alone fails to represent voluminous regions, reducing them to thin lines and losing their proportional weight. For example, in Figure~\ref{fig:hidden_stru_form} (bottom-left), the thick body of the `green cup' becomes a mere round outline. 
To preserve this spatial presence, we augment the skeleton with region-based cues by applying morphological erosion to the original filled shapes. These eroded residuals serve as shrunken proxies for the original masses. The final hidden backbone is defined as the union of the skeleton and these residual areas, capturing both connectivity and proportional volume.

This hidden structure serves as a \lm{unified} profile for both the input target and reference images. It provides a structural-aware intermediate representation without binding to specific contours. \lm{Our experiments demonstrate that this hidden-structure design effectively balances structural preservation with the flexibility required for geometric synthesis during image translation.}

\begin{figure}[!t]
  \centering    \includegraphics[width=1.\linewidth]{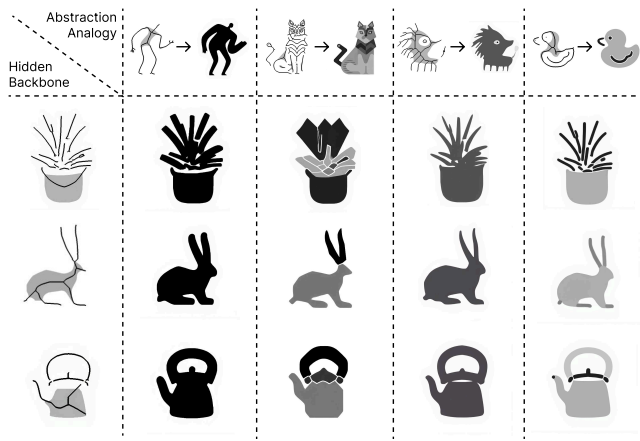}
    \caption{Examples of Abstraction Proxies Inferred by A-VATs: for the same hidden backbone in each row, distinct abstraction proxies are generated by different A-VATs which are trained on different reference respectively.}  
  \label{fig:avat_example}
\end{figure}

\subsection{\textit{Backbone $\rightarrow$ Proxy} Analogy}

To enable analogy-based structural abstraction, we first curate a set of \textit{Hidden Backbone $\rightarrow$ Abstraction Proxy} (shortened as \textit{Backbone $\rightarrow$ Proxy}) paired images from the reference. As shown at the top of Figure~\ref{fig:avat_example}, for each exemplar in the reference set, the pair consists of its hidden structure on the left (i.e., $\text{Backbone}_r$) and its corresponding abstraction proxy on the right (i.e., $\text{Proxy}_r$). The $\text{Proxy}_r$ is obtained by vectorizing the exemplar using the method in Section~\ref{subsec:hidden_struct}. The proxy is rendered in grayscale to eliminate color cues, encouraging the model to learn abstraction logic from the reference structures rather than from appearance correlations. 

Then a training sample for A-VAT is composed of two exemplars $r1$ and $r2$, to show the structural analogy from backbone to proxy (i.e., \textit{$\text{Backbone}_{r1} \rightarrow \text{Proxy}_{r1} :: \text{Backbone}_{r2} \rightarrow \text{Proxy}_{r2}$}). Learning only from a small training set of 5 to 20 samples, the model learns the specific analogy of how to transform a bare structure into that exemplar's abstracted rendition. During inference, this learned analogy A-VAT is applied to an unseen hidden structure (\textit{$\text{Backbone}_t$}). As shown on the right of Figure~\ref{fig:avat_example}, when conditioned on different A-VATs, the same input structure yields distinct proxies (\textit{$\text{Proxy}_t$}).




%% file: tex/32_VisualStyle.tex
\section{Visual Stylization}
\label{sec:stylization}

The Visual Stylization stage renders an abstraction proxy into a final stylized output image (i.e., \textit{AiS Output}) that reflects the visual appearance of the reference style.
This stage does not introduce a new modeling paradigm.
Instead, it reuses the same Visual Analogy Transfer (VAT) mechanism and presents the S-VAT, which accomplishes the following visual analogy:

{\setlength{\abovedisplayskip}{5pt}
\setlength{\belowdisplayskip}{5pt}
\begin{equation*}
    \text{Proxy}_r \rightarrow \text{AiS Output}_r  ::  \text{Proxy}_t \rightarrow \text{AiS Output}_t
\end{equation*}}

where the S-VAT learns to transform abstract representations to stylized outputs from reference examples and uses this mapping to transfer the rendering behavior to the target abstraction proxy.

To train the S-VAT, a small training dataset (5 to 20 samples) is curated. Each training sample shows the analogy from proxy to AiS output by two reference exemplars $r1$ and $r2$, i.e., \textit{$\text{Proxy}_{r1} \rightarrow \text{AiS Output}_{r1}::  \text{Proxy}_{r2} \rightarrow \text{AiS Output}_{r2}$}. In the training sample, two exemplars $r1$ and $r2$ are stacked up and down.
This formulation of the S-VAT mirrors that of the abstraction stage, while operating in a distinct visual domain focused on appearance rather than structure. 


%% file: tex/4_VAT.tex
\section{Visual Analogy Transfer (VAT)}
\label{sec:vat}

We present \textit{Visual Analogy Transfer (VAT)} as a general mechanism for learning and applying transformations between visual representations in image space.
Rather than being tied to a specific task or representation, VAT formulates a transformation as a visual analogy problem ($A \rightarrow A' :: B \rightarrow B'$): given an example pair that demonstrates how one visual representation $A$ is transformed into another $A'$, the same transformation can be applied to a new input $B$ and inferred $B'$. At a high level, VAT follows the classical notion of visual analogy\cite{hertzmann2001}. Unlike traditional approaches, which rely on handcrafted geometric correspondences, symbolic structures, or explicit rules, VAT does not assume any of these pre‑defined constraints. \lm{Instead, it leverages the expressive priors of pretrained image generation models to automatically learn and apply complex, style‑dependent transformations directly in image space.}

\begin{figure}[!htb]
  \centering
    \includegraphics[width=1.\linewidth]{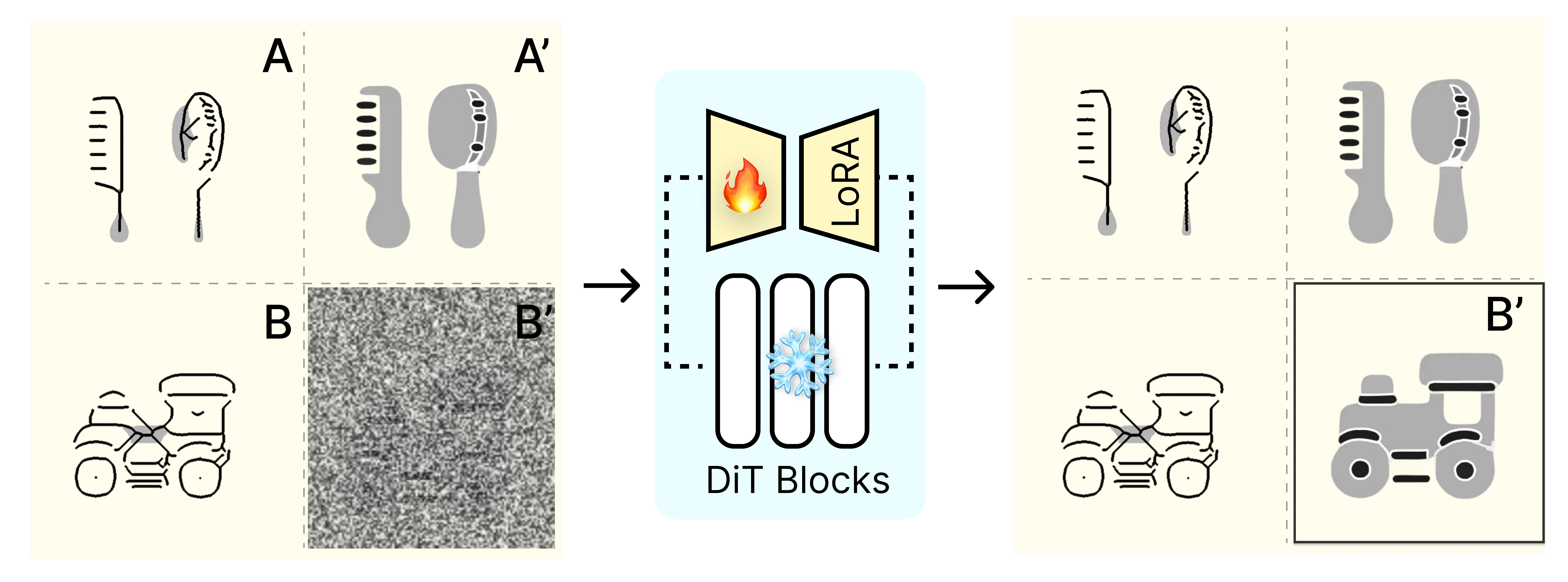}
    \caption{Visual Analogy Transfer (VAT) realized with DiT and LoRA: A visual transformation is expressed as a $2 \times 2$ analogy, where a reference pair ($A \rightarrow A'$) defines a relation that is applied to a new input ($B$) by predicting the missing panel ($B'$).
VAT is implemented using a Diffusion Transformer (DiT) conditioned on the visible panels, with a lightweight LoRA used to adapt the model to the analogy relation.}
  \label{fig:VAT}
\end{figure}

\lm{While our Visual Analogy Transfer (VAT) draws inspiration from the broader paradigm of visual in-context learning~\cite{chen2025edittransfer, vsubrtova2023diffusion, gu2024analogist, yang2023imagebrush}, it diverges fundamentally in its operational objective. Where existing methods largely rely on semantic analogy mechanisms, focusing on transferring high-level relational changes and editing operations. In contrast, VAT addresses the specific challenge of style-driven transformation. It focuses on cases where shapes must be reinterpreted and morphed in ways that extend beyond traditional semantic labels, operating instead on direct visual logic.} As illustrated on the left of Figure~\ref{fig:VAT}, we instantiate VAT using a $2 \times 2$ layout image. The top row presents a reference pair illustrating a transformation from visual representation $\mathcal{A}$ to another $\mathcal{A'}$. 
The bottom-left panel contains a new input representation $\mathcal{B}$, while the bottom-right panel for $\mathcal{B'}$, initially masked, is generated by the model.
By conditioning on the three visible panels, the model learns to infer $B'$ from $B$ via the transformation that relates the reference pair.

\begin{figure*}[!t]
  \centering    
  \includegraphics[width=1.\linewidth]{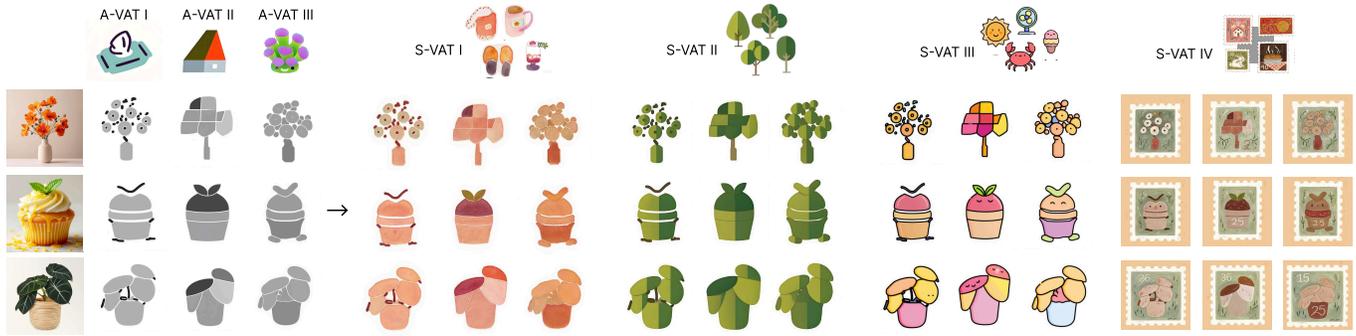}
    \caption{Mix of Three Different A-VATs with Four S-VATs: giving the same S-VAT, the generated outputs share visual appearance (color, texture) but exhibit distinct geometry, demonstrating independent control of style and structure.}
  \label{fig:mixup}
\end{figure*}

To realize VAT in practice, we fine-tune a lightweight low-rank adapter (LoRA)~\cite{hu2022lora} on an image inpainting diffusion transformer model FLUX.1-Fill-dev.
The base model remains frozen, while the LoRA adapts the model to internalize the analogy-specific transformation.
Within Abstraction in Style, VAT is employed in two distinct contexts.
In the Structural Abstraction stage (Section \ref{sec:structural_abstraction}), A-VAT is to fine-tune with a LoRA on a small set of \textit{Backbone $\rightarrow$ Proxy} $2 \times 2$ grid images. Similarly, S-VAT is fine-tuned with LoRA on several \textit{Proxy $\rightarrow$ AiS Output} $2 \times 2$ grid images. For each, the number of $2 \times 2$ grid training images is around 5 to 20, composed from 10 to 40 examplars.


By decoupling the mechanism of analogy learning from the specific representations it operates on, VAT provides a flexible and reusable tool for image-space transformation.
This generality allows Abstraction in Style to be realized through a unified mechanism, while preserving a clear conceptual separation between abstraction and stylization.



%% file: tex/5_experiments.tex
\section{Results and Experiments}

Figure~\ref{fig:two_stage_examples} shows the results of four target images rendered in five distinct \lm{illustrative} styles by AiS. As seen, given the same target \lm{and backbone} for each row, different abstraction proxies are inferred by the specific trained A-VATs. In Figure~\ref{fig:one_for_all_short}, we show more challenging abstracted cases. In each column, the same target image is generated into different art styles. Figure~\ref{fig:three_mixup} demonstrates visual designs that intermix reference exemplars with AiS-generated results to test for stylistic consistency. As shown, the generated results blend seamlessly with the exemplars, maintaining visual harmony with nearly imperceptible differences.

The disentanglement of A-VAT and S-VAT provides controllability in the stylized image generation. It allows for mixing up A-VAT and S-VAT to create new designs. In Figure~\ref{fig:mixup}, we combine three distinct A-VATs with four S-VATs. As shown, the A-VAT decisively determines the geometric layout and structural essence of the result, while the S-VAT consistently provides the stylistic rendering. This systematic mixing validates the robustness of our disentanglement and highlights its potential for controllable generative design. 



\setlength{\columnsep}{5pt} 

\begin{figure}[!htb]
  \centering    \includegraphics[width=.7\linewidth]{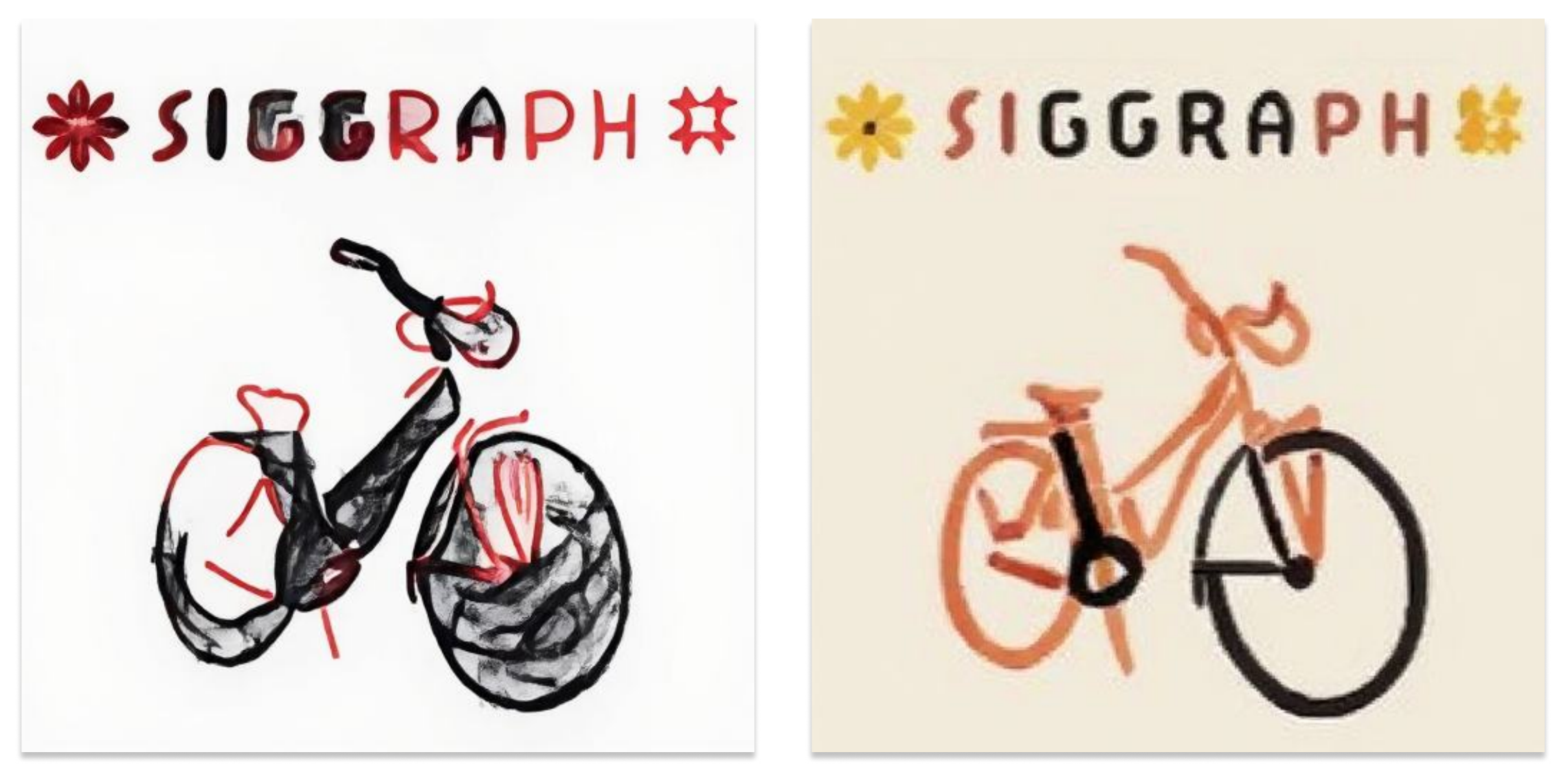}
    \caption{Generated examples in which textual elements are stylized with the primary visual content.}
  \label{fig:small_example}
\end{figure}

AiS can stylize auxiliary visual elements beyond the main object. 
As illustrated in Figure~\ref{fig:small_example}, textual elements rendered in custom fonts are stylized alongside the primary content, indicating that the abstraction and stylization process applies holistically to the entire image rather than being limited to object-centric regions.

\subsection{Ablation Study} 

\begin{figure}[!htb]
  \centering    \includegraphics[width=1.\linewidth]{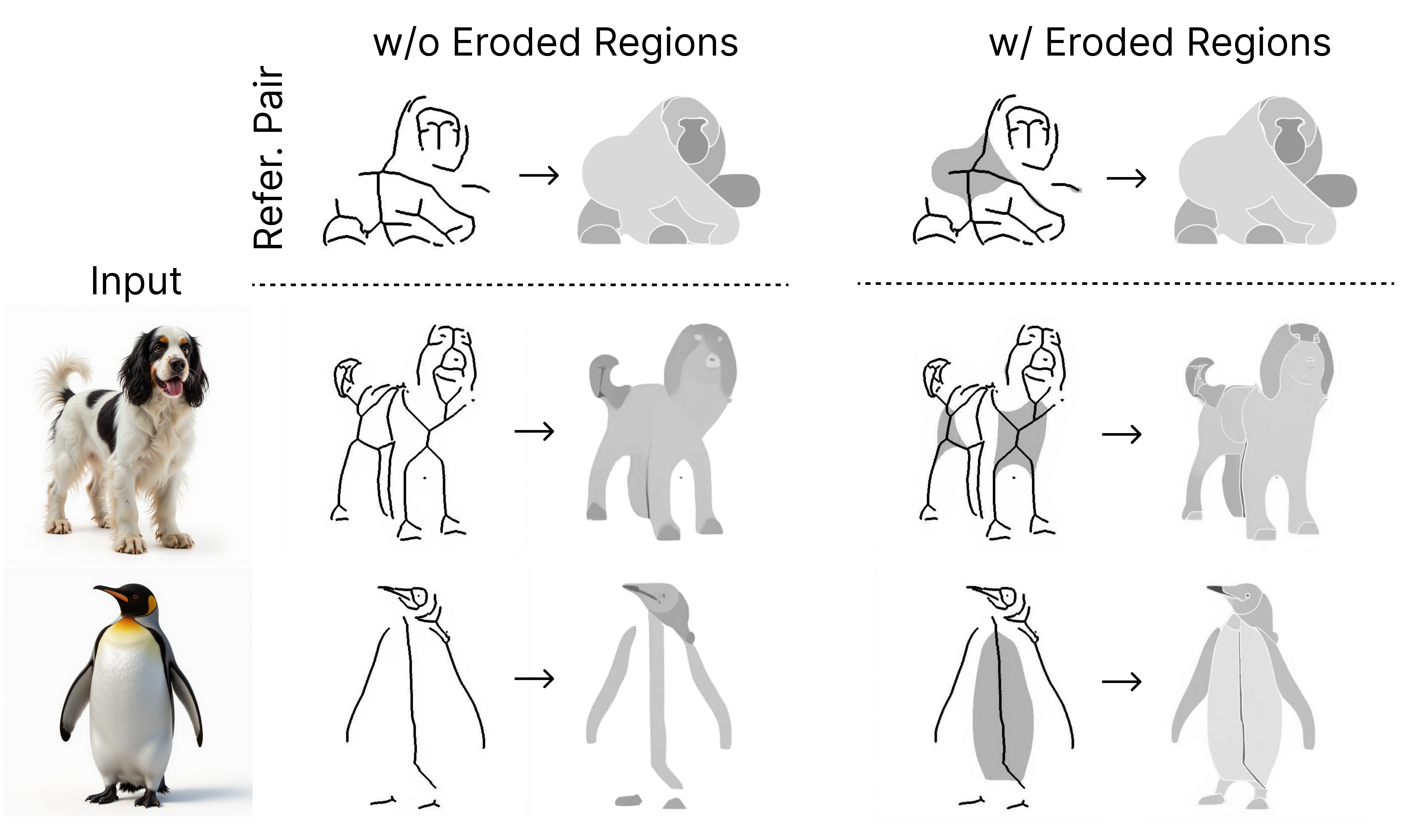}
    \caption{Ablation of the Eroded Region Cues in Hidden Backbone: skeleton-only representation (i.e., without eroded regions) can produce a geometrically degraded abstraction proxy.}
  \label{fig:ablation_hidden}
\end{figure}

\paragraph{Ablation on the Design of Hidden Backbone} 
We ablate the eroded-region cues from the hidden backbone's representation. As shown in Figure~\ref{fig:ablation_hidden}, training the A-VAT module without these cues produces geometrically degraded outputs in the abstraction proxy, impairing structural integrity in regions such as the dog's chest and the penguin's body.

\begin{figure}[!htb]
  \centering    \includegraphics[width=1.\linewidth]{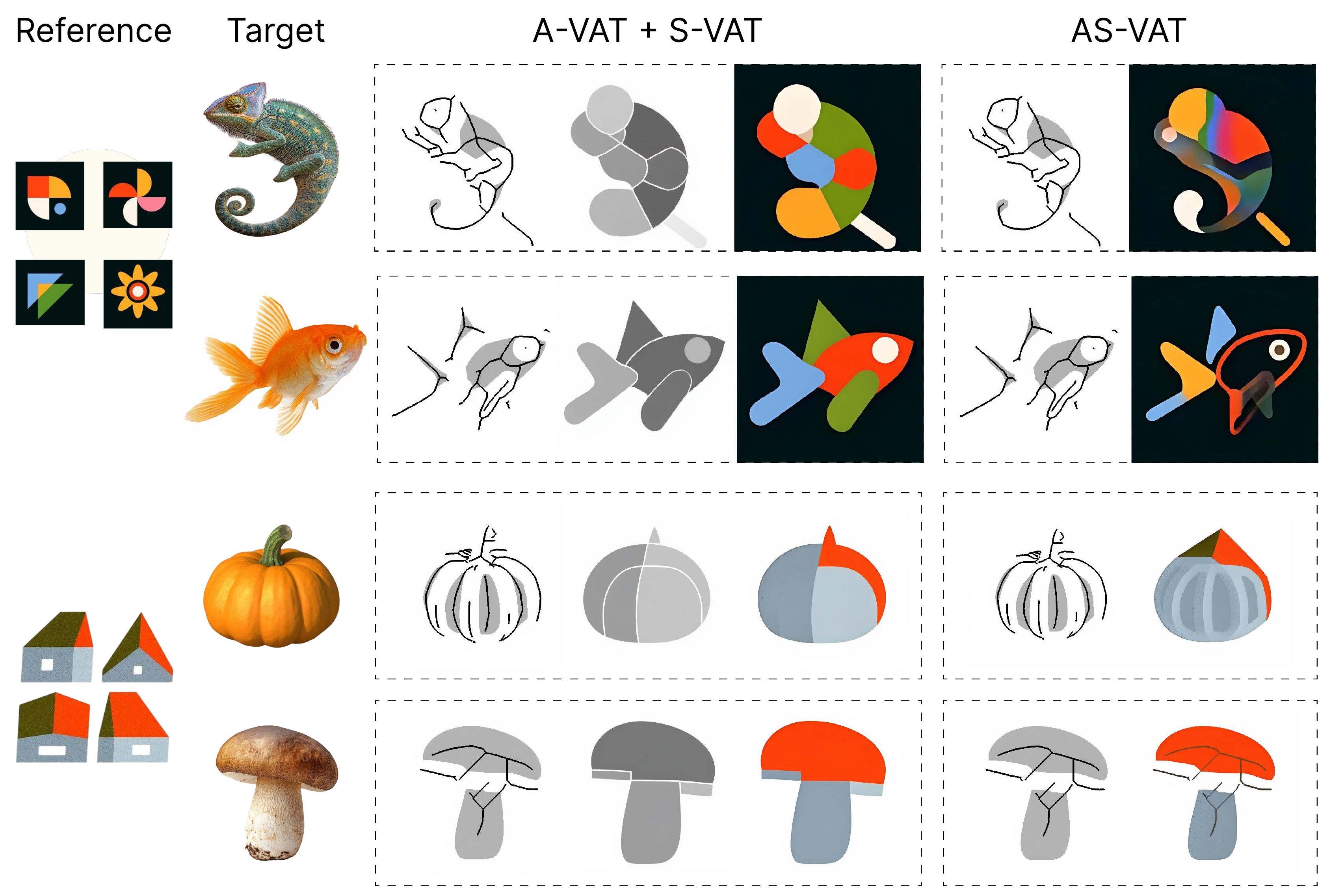}
    \caption{Comparison of the two-stage (A-VAT + S-VAT) and a distilled single-stage variant (AS-VAT). \lm{The single-stage AS-VAT fails to synthesize coherent results, yielding visual artifacts such as "holes," color bleeding, and rigid adherence to input geometry.}}
  \label{fig:ablation_geo}
\end{figure}

\paragraph{Ablation on the VAT Design}
We ablate the two-stage VAT design (A-VAT and S-VAT) by comparing it with a distilled single-stage variant, AS-VAT, which directly maps the hidden backbone to the final output (\textit{Backbone $\rightarrow$ Output}), bypassing the abstraction proxy.
AS-VAT is trained using the same image scales as the full framework (averagely ten $2\times2$ training images).
As shown in Figure~\ref{fig:ablation_geo}, only the two-stage pipeline with an explicit abstraction proxy produces outputs that are both structurally coherent and well aligned with the reference style.
\lm{The single-stage design tends to adhere to the input geometry, fails to properly reinterpret structure, and causes style inconsistent with the reference style.}
Beyond quality, this decoupled design also enables post-hoc controllability, as demonstrated in Figure~\ref{fig:mixup}, independently trained A-VAT and S-VAT modules can be recombined to synthesize hybrid styles.

\begin{figure*}[!htb]
  \centering    \includegraphics[width=1.\linewidth]{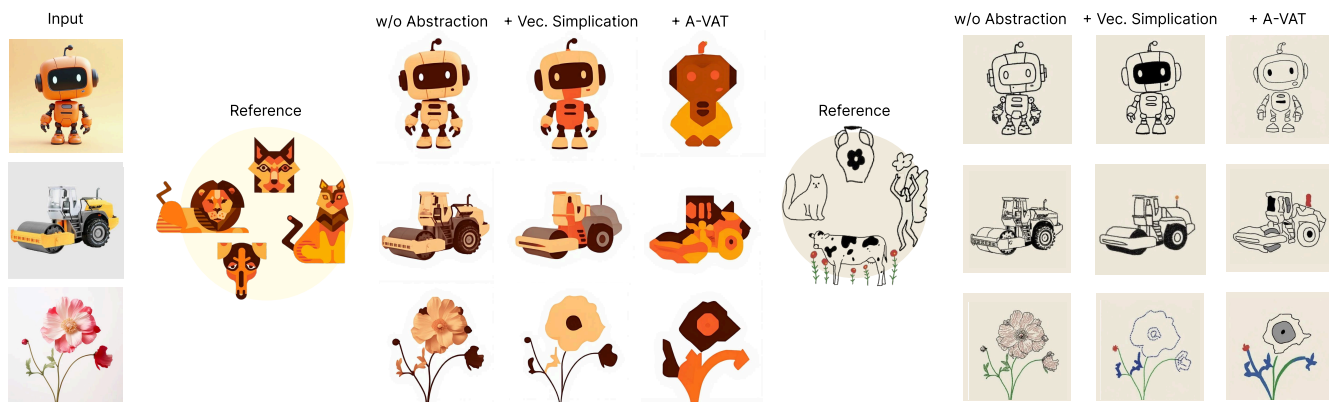}
    \caption{Ablation of the Abstraction Stage: the two ablated framework variants, \textit{without Abstraction} and \textit{with only Vector Simplification}, fail to generalize visual logic. Their outputs exhibit geometry that strictly adheres to the original input target. 
    }
  \label{fig:ablation_abs}
\end{figure*}

\paragraph{Ablation on the Abstraction Stage} We investigate the impace of the abstraction stage via two variants: (1) \textit{w/o Abstraction}, where the original image serves as the proxy for S-VAT. (2) \textit{$+$ Vec. Simplification}, where vector paths are used directly as the proxy, by passing the A-VAT module. As shown in Figure~\ref{fig:ablation_abs}, only the full abstraction pipeline successfully captures the abstract visual logic of the reference examples. In contrast, both ablated variants merely conform to the original input geometry, failing to reinterpret its underlying structure. \lm{Quantitative comparison with these ablations is shown in Table~\ref{tab:quantitative}, with metrics provided in Section~\ref{subsec:compare}.}

\subsection{Comparisons with Baselines}
\label{subsec:compare}

Figure~\ref{fig:comparison_short} provides a qualitative comparison against state-of-the-art methods, including direct style transfer (StyleID~\cite{chung2024style}, StyleAlign~\cite{hertz2024stylealigned}, Attention Distillation~\cite{AttDistill25}), LoRA-based adaptation (B-LoRA~\cite{frenkel2024blora}, ZipLoRA~\cite{shah2024ziplora}, K-LoRA~\cite{ouyang2025k}, and the Nano Banana Pro public model~\cite{team2023gemini}. AiS demonstrates superior alignment with both the reference style’s visual characteristics and its structural abstraction. This yields outputs with more coherent stylistic expression and consistently appropriate geometric reinterpretation, whereas other methods (e.g., Nano Banana) largely preserve the original target geometry, limiting structural adaptation.

\begin{table}[!htb]
    \centering
    \footnotesize
    \renewcommand{\arraystretch}{0.85}
    \caption{\lm{Quantitative comparison of baseline methods and ablations.}}
    \label{tab:quantitative}
    \begin{tabular}{l ccccc}
        \toprule
        \textbf{Metrics} & StyleID & A.D. & StyleA. & B-LoRA & ZipLoRA \\
        \midrule
        CSD       & 0.36 & 0.54 & 0.22 & 0.37 & 0.34 \\
        Style FID & 529.05 & 431.12 & 511.59 & 495.20 & 547.73 \\
        CLIP-I    & 0.62 & 0.69 & 0.61 & 0.63 & 0.66 \\
        \midrule[\heavyrulewidth]
        & \multirow{2}{*}{K-LoRA} & \multirow{2}{*}{Nano Ban.} & \multicolumn{3}{c}{\textit{AiS (Ablation Study)}} \\ 
        \cmidrule(lr){4-6}
        & & & w/o Abs. & + Vec. Simpl. & \textbf{AiS (Full)} \\
        \midrule
        CSD       & 0.25 & 0.56 & 0.46 & 0.66 & \textbf{0.75 $\uparrow$} \\
        Style FID & 517.42   & 451.17   & 373.03 & 324.84 & \textbf{260.43 $\downarrow$} \\
        CLIP-I    & 0.60   & 0.68   & 0.64 & 0.71 & \textbf{0.76 $\uparrow$} \\
        \bottomrule
    \end{tabular}
\end{table}

\paragraph{Quantitative Analysis} We construct a testing dataset with 10 design styles from \textit{Pinterest} and for each two target images randomly generated by FLUX.1-dev, covering wide topics e.g., animals, buildings, food, objects, yielding 20 test samples. \lm{We adopt three metrics to evaluate style similarity between the generated results and the reference, including Contrastive Style Descriptors (CSD)~\cite{somepalli2024measuringstylesimilaritydiffusion}, CLIP-I (image to image CLIP score)~\cite{radford2021learningtransferablevisualmodels} and Style FID~\cite{heusel2017gans}.
As shown in Table~\ref{tab:quantitative}, our method demonstrates superior performance.}

\paragraph{User Study} We conducted a blind comparative evaluation across the 20 test cases. 61 Users were asked to select the result that best matched the reference style while preserving the target’s essence. Our method was the most preferred, selected in 50\% of trials, followed by Nano Banana (35\%) and Attention Distillation (15\%). All 20 test cases are available in the supplementary material. \lm{Feedback from two veteran designers (10+ years exp.) confirmed that our method excels at capturing nuanced styles through structural variation, whereas baselines remain "over-rigidly" tied to input geometry. Experts also highlighted that decoupling abstraction from appearance significantly improves both stylistic fidelity and user controllability.}

%% file: tex/discussion.tex
\section{Conclusion, Limitations, and Future Work}

We introduced \emph{Abstraction in Style} (AiS), a generative framework that treats structural abstraction as an explicit and transferable component of stylized image generation. By decomposing stylization into two sequential stages, structural abstraction followed by visual stylization, \lm{AiS enables the image style transfer that move beyond standard visual traits such as color and texture to encompass underlying geometric behaviors.} Our concrete realization of AiS leverages an intermediate abstraction proxy and a shared image-space analogy mechanism to operationalize this separation. Empirically, we show that explicitly modeling abstraction leads to more coherent structural reinterpretations, improved stylistic fidelity, and greater controllability compared to conventional single-stage style transfer approaches, particularly for \lm{highly abstract or illustrative styles}. 


\setlength{\columnsep}{5pt} 


In its current form, our implementation represents only one instantiation of the broader AiS paradigm. \lm{The current hidden backbone is limited by a lack of semantic awareness. 
In some cases, the abstraction stage fails to distinguish between the skeletons of eroded versus naturally thin regions, leading to visual artifacts. More dedicated abstraction proxy shall be introduced in the future. Looking ahead, we envision richer abstraction representations and learning mechanisms, allowing models to selectively reshape structure to deal with creative abstraction behaviors that involve strong semantic distortion, exaggeration, or deliberate disproportions.}


\lm{Beyond this specific implementation, our work points toward a larger shift in how we view stylization.}
Our findings suggest that abstraction and appearance play fundamentally different roles in stylized generation. While it may be tempting to pursue a unified, single-stage formulation, our results indicate that explicitly separating abstraction from stylization provides both conceptual clarity and practical benefits.

%% file: tex/figures_only.tex

\begin{figure*}[!htb]
  \centering    \includegraphics[width=1.\linewidth]{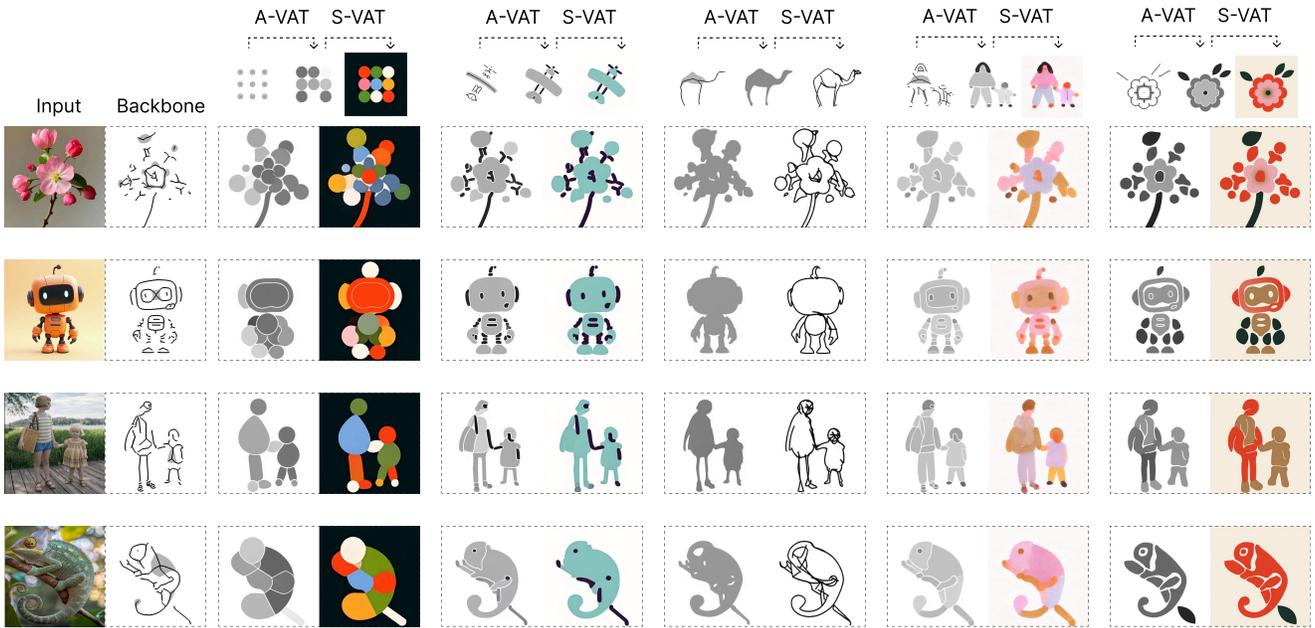}
    \caption{Generated examples:  For the target image in each row, its style-agnostic hidden backbone is first generated. Then, five stylized outputs are produced, one for each reference style (columns). For each style, the system uses a trained A-VAT to generate an abstraction proxy (left grayscale image in each pair), which is then stylized via the corresponding S-VAT to create the final result (right image in the pair). }    
  \label{fig:two_stage_examples}
\end{figure*}

\begin{figure*}[!htb]
  \centering    \includegraphics[width=1.\linewidth]{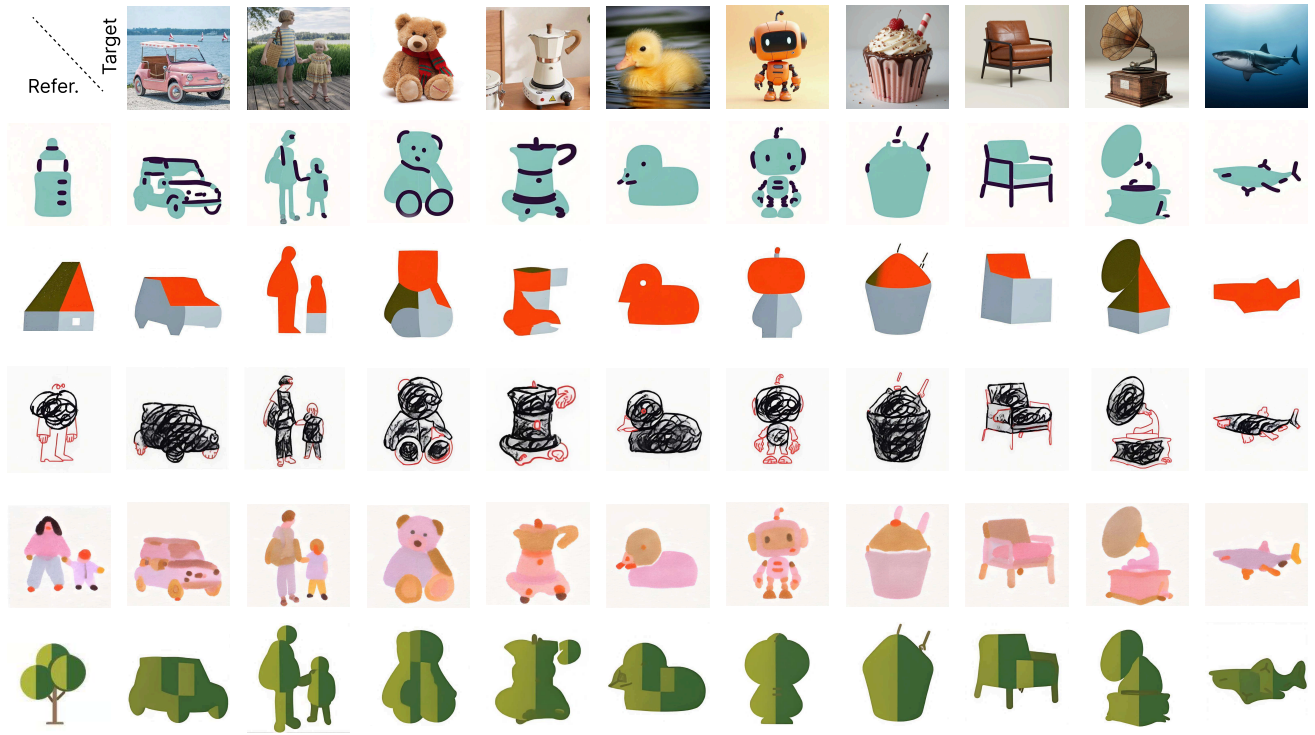}
    \caption{Generated examples: Each column corresponds to a different target image. Each row corresponds to a different reference style (with one exemplar shown per style), which is applied to all targets in that row. The full set for each style comprises 5 to 10 exemplars, available in the supplementary material.}    
  \label{fig:one_for_all_short}
\end{figure*}

\begin{figure*}[]
  \centering    \includegraphics[width=1.\linewidth]{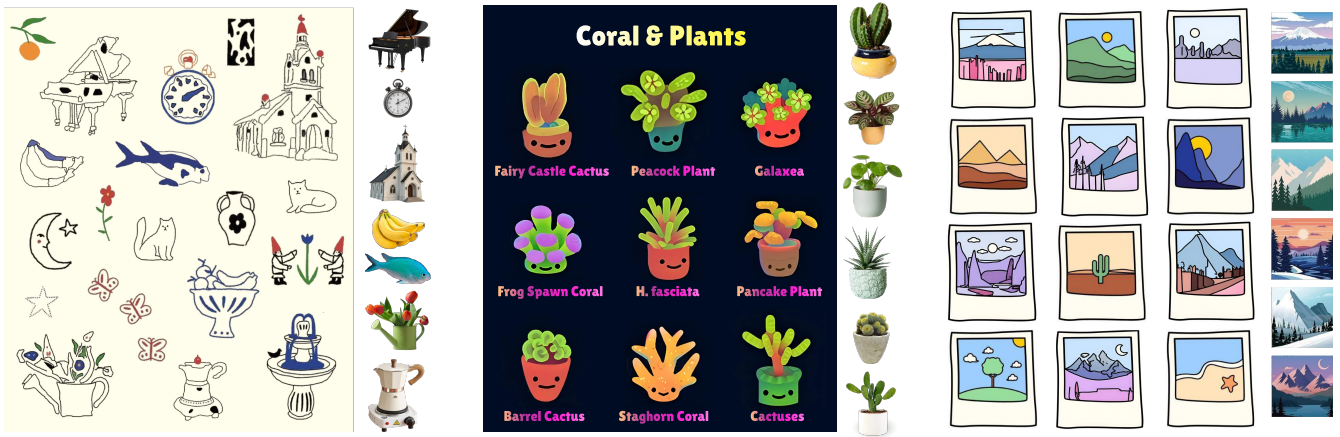}
    \caption{Design examples mixing up the reference exemplars and the results generated by AiS: on the side are the original images from which the results are generated. The original images are ordered according to the result positions in the example, from top to bottom, left to right.}    
  \label{fig:three_mixup}
\end{figure*}

\begin{figure*}[!htb]
  \centering    \includegraphics[width=1.\linewidth]{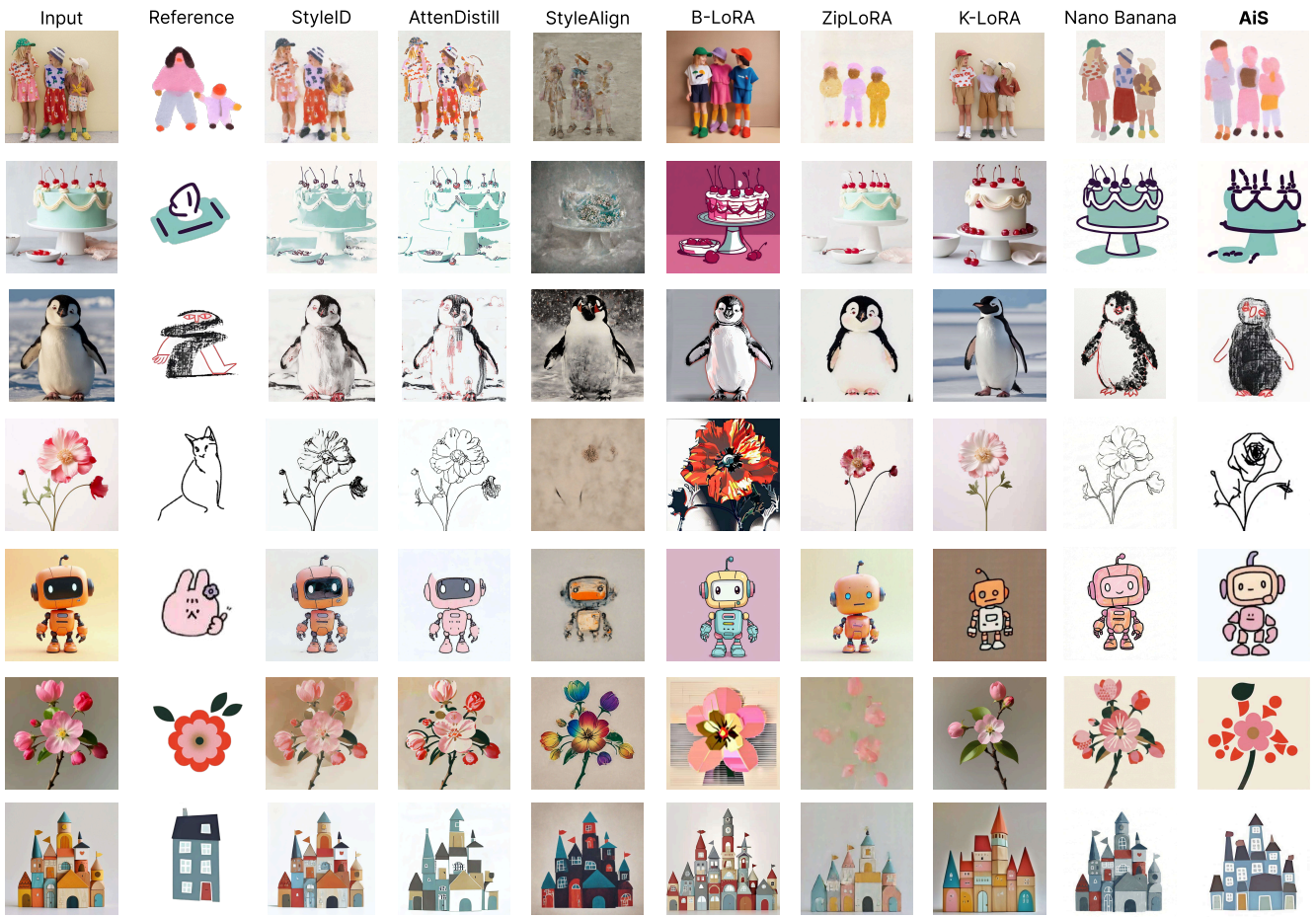}
    \caption{Qualitative comparison with baseline methods: existing methods struggle with abstract styles and over-rigidly preserve input structure, producing unsatisfactory results. Our method excels at capturing nuanced artistic styles while maintaining natural structural variations. A larger version of this figure, including additional examples, is provided in the supplementary material.}
  \label{fig:comparison_short}
\end{figure*}

%% file: tex/supp_content.tex
\section{Implementation Details}

\paragraph{Hidden Backbone Detection} During vectorization, vector shapes are sorted into back-to-front layers. By default, shapes related to the centered objects in the three rearmost layers are selected and jointly rendered into a binary hidden-backbone image, though manual removal of fragmented shapes is occasionally required. Each shape is filled in black and outlined with a thin white stroke (2 pixels), preserving distinctness under overlap. The resulting binary image is then processed using \texttt{skeletonize} function followed by \texttt{binary\_erosion} with a disk (radius set to 25 pixel) in scikit-image's morphology module, to get the hidden backbone image.

\paragraph{Style Reference \& Exemplars}

\begin{figure}[!htb]
  \centering    \includegraphics[width=1.\linewidth]{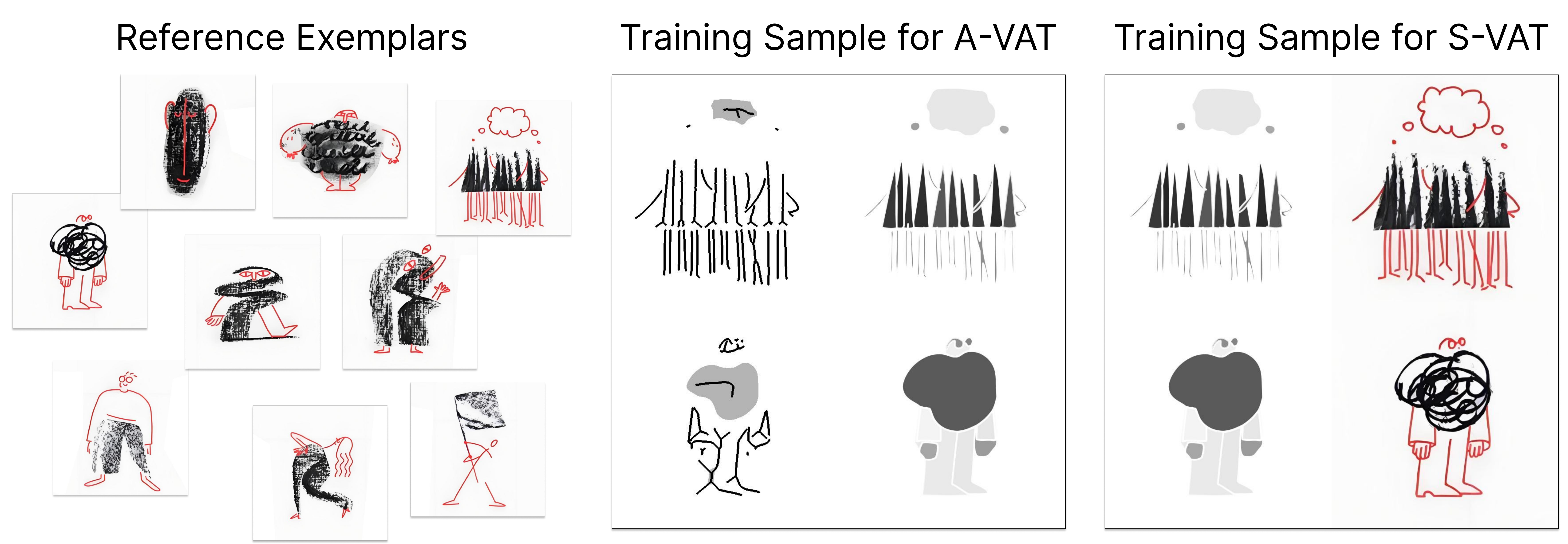}
    \caption{Reference exemplars and $2 \times 2$ image training samples for A-VAT (\textit{Backbone $\rightarrow$ Proxy}) and S-VAT (\textit{Proxy $\rightarrow$ Output).}}
  \label{fig:reference_example}
\end{figure}

Each reference style is defined by \( N \) exemplars, typically ranging from 5 to 40. From these exemplars, we compose \( 2 \times 2 \) grid images by stacking two examplars without repetition, yielding a small training set of 5 to 20 images per style. Figure~\ref{fig:reference_example} shows a style reference with 9 exemplars along with the corresponding training image samples for A-VAT and S-VAT. The full set of exemplars for sampled styles used in our work is shown in Figure~\ref{fig:all_styles}. All reference images in this work are sourced from Pinterest \textcopyright{} original authors.

\paragraph{VAT Training \& Training Data} 
For each reference style, we train two separate LoRA adapters on the \texttt{FLUX.1-Fill-dev} model: one for the A-VAT and another for the S-VAT. Each adapter is trained for 1,000 steps with a rank of 16. Training one LoRA adapter takes approximately one hour on a single A100 (80GB) GPU at \(1024 \times 1024\) resolution. 

For each VAT, we fine-tuned \texttt{FLUX.1-Fill.Dev} on 5 to 20 $2 \times 2$ images. The right of Figure~\ref{fig:reference_example} shows the training $2 \times 2$ image samples for A-VAT and S-VAT respectively. Each image is paired with a corresponding text prompt. We employed the following template for these prompts. 

\begin{tcolorbox}[
    colback=blue!3,
    colframe=blue!50!black,
    boxrule=0.2pt,       
    arc=2pt,              
    left=10pt,
    right=10pt,
    top=10pt,
    bottom=10pt,
    fontupper=\sffamily\small
]
This is a four-panel image on a uniform solid-color background, hand-drawn in style, with the subjects highlighted and kept as simple as possible:

\medskip

{\ttfamily [TOP-LEFT]}: Image of the structure of a subject. \\
{\ttfamily [TOP-RIGHT]}: An edited version of the {\ttfamily [TOP-LEFT]} image, transformed to {\itshape [styvec]} style. \\
{\ttfamily [BOTTOM-LEFT]}: Skeleton or structural image of another subject. \\
{\ttfamily [BOTTOM-RIGHT]}: An edited version of the {\ttfamily [BOTTOM-LEFT]} image, applying the same style transformation as used in {\ttfamily [TOP-RIGHT]}.
\end{tcolorbox}

This template does not provide specific textual descriptions for the depicted subjects; they are generically labeled as "a subject" or "another subject."




\section{Additional Ablation Studies}

\begin{figure}[!htb]
  \centering    \includegraphics[width=.9\linewidth]{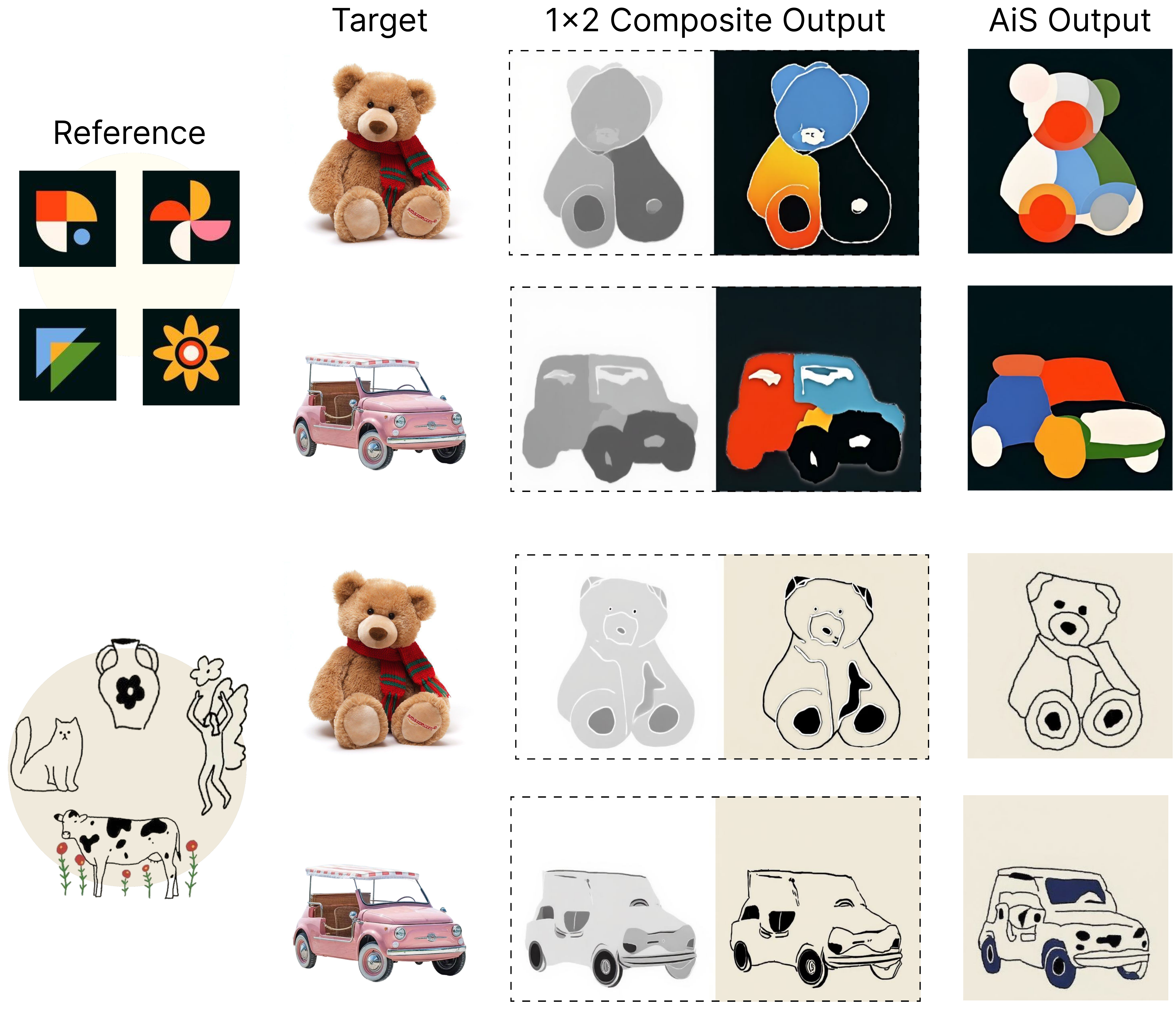}
    \caption{Ablation of the $2 \times 2$ Analogy Layout Design: training S-VAT with $1\times2$ image composite fails to learn the effective style transferring.}
  \label{fig:ablation_layout}
\end{figure}

\paragraph{Ablation on $2\times2$ Image Composite Layout} We evaluate the necessity of the standard $2\times2$ image composite (2 rows, 2 columns) by ablating to a simpler $1 \times 2$ layout (1 rows, 2 columns) during training. Results in Figure~\ref{fig:ablation_layout} show that training with only a $1\times2$ composite is insufficient. The resulting S-VAT model fails to learn effective style alignment, underperforming the model trained on the full $2\times2$ layout.

\begin{figure}[!h]
  \centering    \includegraphics[width=.85\linewidth]{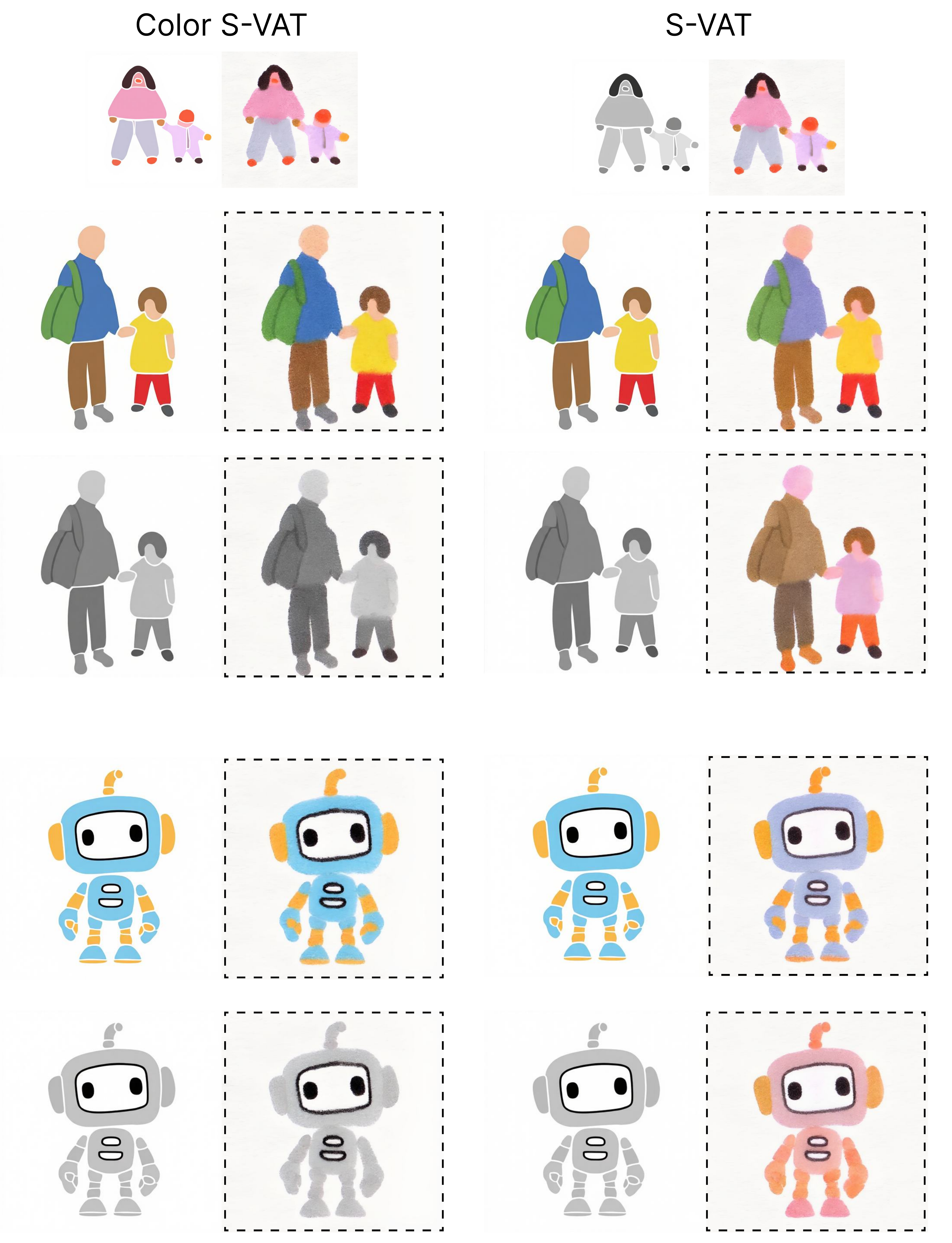}
    \caption{Ablation of the Grayscale Proxy Design: In Color S-VAT (color proxy $\rightarrow$ output), the color from the proxy is effectively transferred to the output. In contrast, S-VAT (grayscale proxy $\rightarrow$ output) infers new colors for the proxy that align with the reference.}
  \label{fig:color_ablation}
\end{figure}

\paragraph{Ablation on the Color of Abstraction Proxy} We ablate the grayscale design of the abstraction proxy in S-VAT by comparing it with a version that retains color, referred to as Color S-VAT (using a Color Proxy → AiS Output pipeline). Color S-VAT was trained using the same image scale as standard S-VAT (Proxy → AiS Output). Results are shown in Figure~\ref{fig:color_ablation}. In Color S-VAT, the target’s color is effectively passed to the output because the model directly observes color consistency from the proxy. In contrast, with the grayscale proxy, the model must infer new colors, a capability learned intrinsically from the reference exemplars. Since the goal of AiS is to align the target's style (including color) to the reference, we opt for a grayscale proxy.


\section{Evaluation Details}



\paragraph{Baseline Settings} To make a fair comparison with baseline methods, we tested on StyleAlign~\cite{hertz2024stylealigned} with ControlNet~\cite{zhang2023adding} to input the condition of the image depth. For Attention Distillation~\cite{AttDistill25}, StyleID~\cite{chung2024style} and other LoRA-based methods, they were tested using their official implementations with default hyperparameters, unless otherwise noted. 
To standardize the input, all target images were first resized to $1024 \times 1024$ resolution before being fed into the models, regardless of the original input size. This consistent evaluation set-up allows for a direct and objective comparison across methods in terms of style alignment quality.

For the Nano Banana Pro model, a single input image is first created by placing a reference exemplar to the left of the target image. Then the mode is prompted to generate the stylized output using the following query `\textit{Apply the style of the Reference image (left sub-image) to the Target Image (right sub-image) while preserving its original content}'. For each target image, we generated five outputs and selected the best one. 


\paragraph{Testings for User Study} The 20 testings (10 styles $\times$ 2 subjects) for the user study (Section 7) are shown in Figures~\ref{fig:comparison_eval1} and ~\ref{fig:comparison_eval2}. 

\paragraph{Time Cost} For feed-forward methods, such as StyleAlign, StyleID, achieve near-instantaneous inference in under one minute. Below we report the computational overhead required to generate a single stylized content image using LoRA-based methods. For B-LoRA~\cite{frenkel2024blora}, the total process of generating a single stylized image takes approximately 15 minutes on a single NVIDIA RTX 4090 (24GB) GPU, which includes roughly 7 minutes each for content and style training, followed by less than one minute for inference. ZipLoRA~\cite{shah2024ziplora} requires a significantly higher time cost of approximately 85 minutes on an NVIDIA A100 (40GB) GPU. This duration encompasses 35 minutes for individual content and style training sessions, respectively, plus an additional 14 minutes for the LoRA merging (zipping) optimization phase. For AiS, for each reference style, it takes approximately one hour to train a VAT, therefore two hours for A-VAT and S-VAT in total on a single H100 (80GB) GPU. Once A-VATs and S-VATs are trained, they can be combined and achieve near-instantaneous inference in under one minute.


\section{Additional Results}



Figures~\ref{fig:one_for_all1} and ~\ref{fig:one_for_all2} show more examples. Figure~\ref{fig:comparison} shows more comparisons between our method and baseline methods. Figures~\ref{fig:combo_1} and ~\ref{fig:combo_2} show the stylized sets generated by AiS.

%% file: tex/figure_only_supp.tex
\clearpage

\begin{figure*}[!htb]
  \centering    \includegraphics[width=1.\linewidth]
  {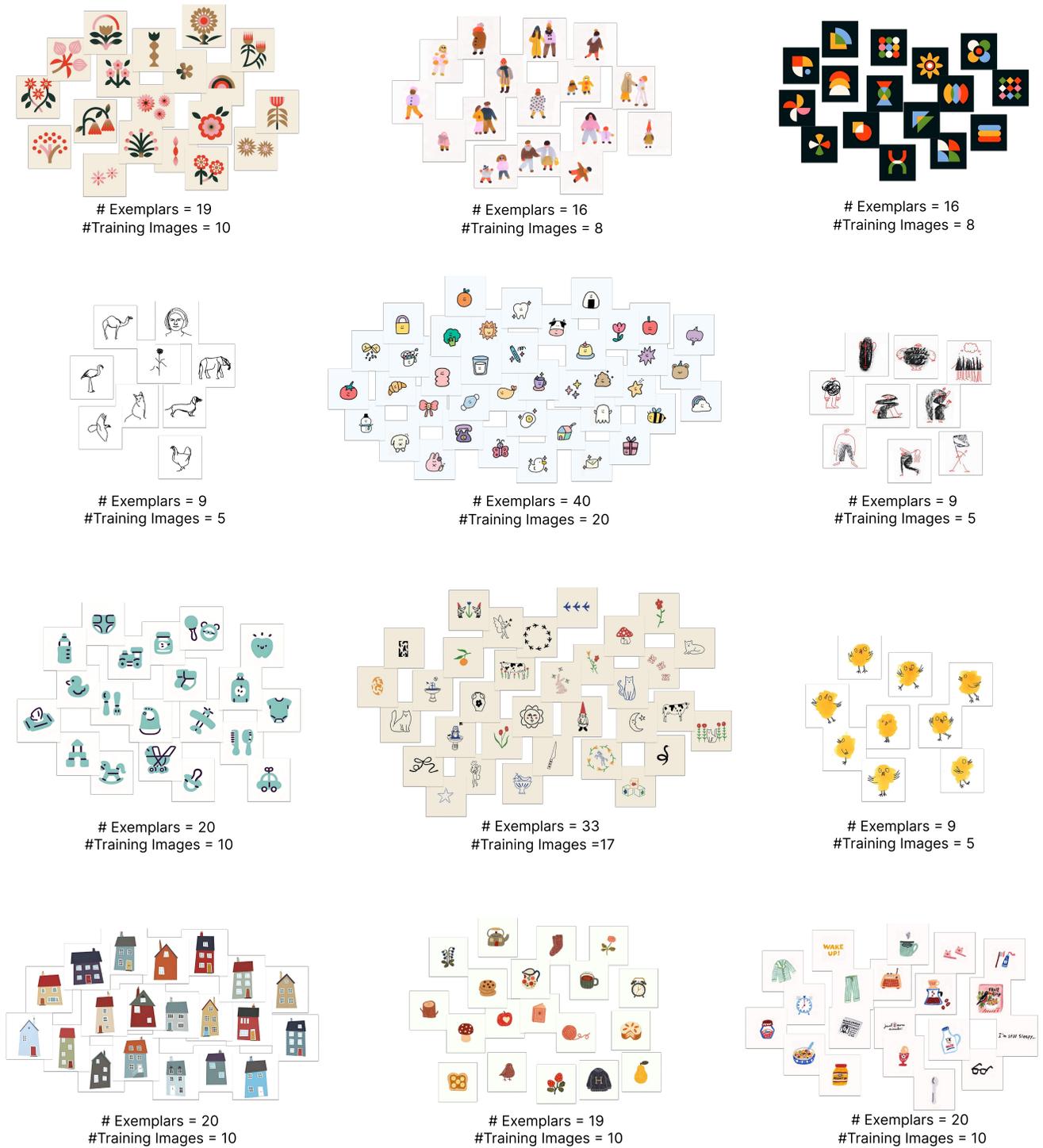}
    \caption{Full sets of exemplars in sampled references: for each reference style, it shows the number of exemplars in the set (\# Exemplars) and the number of training images for VAT (\# Training Images). }
  \label{fig:all_styles}
\end{figure*}

\begin{figure*}[!htb]
  \centering    \includegraphics[width=1.\linewidth]{images_supp/gallery.pdf}
    \caption{Gallery of generated examples (part I).}    
  \label{fig:one_for_all1}
\end{figure*}

\begin{figure*}[!htb]
  \centering    \includegraphics[width=1.\linewidth]{images_supp/gallery_2.pdf}
    \caption{Gallery of generated examples (part II).}    
  \label{fig:one_for_all2}
\end{figure*}

\begin{figure*}[!htb]
  \centering    \includegraphics[width=1.\linewidth]{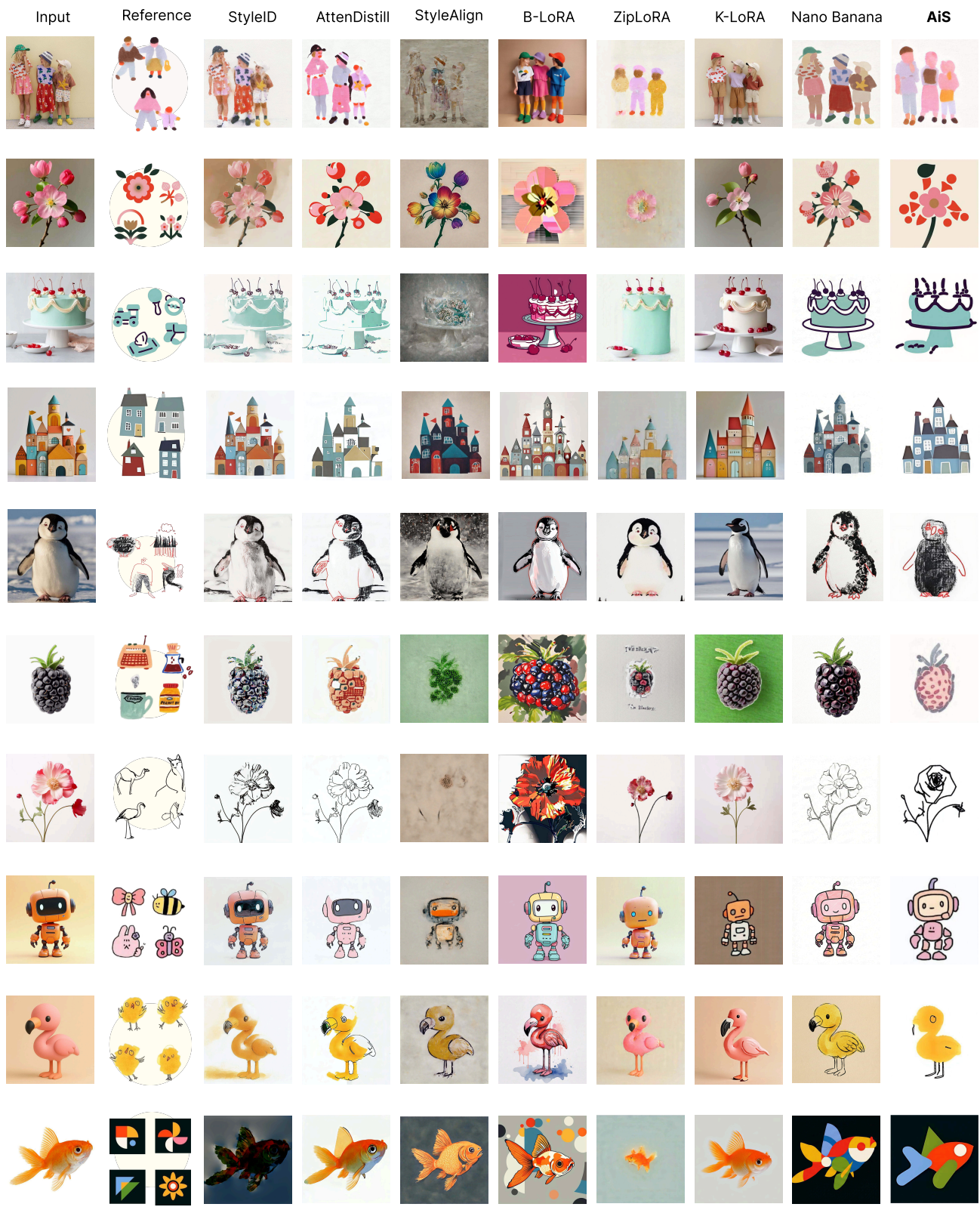}
    \caption{Qualitative comparison with baseline methods.}
  \label{fig:comparison}
\end{figure*}

\begin{figure*}[!htb]
  \centering    \includegraphics[width=1.\linewidth]{images_supp/Comparison_1.pdf}
    \caption{Evaluation tests with baseline methods (part I).}
  \label{fig:comparison_eval1}
\end{figure*}

\begin{figure*}[!htb]
  \centering    \includegraphics[width=1.\linewidth]{images_supp/Comparison_2.pdf}
    \caption{Evaluation tests with baseline methods (part II).}
  \label{fig:comparison_eval2}
\end{figure*}

\begin{figure*}[!htb]
  \centering    \includegraphics[width=1.\linewidth]{images_supp/combo_1.pdf}
    \caption{Generated consistent design sets from diverse inputs (part I).}
  \label{fig:combo_1}
\end{figure*}

\begin{figure*}[!htb]
  \centering    \includegraphics[width=1.\linewidth]{images_supp/combo_2.pdf}
    \caption{Generated consistent design sets from diverse inputs (part II).}
  \label{fig:combo_2}
\end{figure*}